\documentclass[12pt, oneside]{book}
\usepackage{graphicx}
\usepackage{latexsym}
\usepackage{float}
\usepackage{enumerate}
\usepackage{pdfpages}
\usepackage{amsmath}
\usepackage{amsfonts}
\usepackage{amssymb}
\usepackage[numbers, sort]{natbib}
\usepackage{array}
\usepackage{booktabs}
\usepackage[font=small, labelfont=bf, tableposition=top]{caption}
\usepackage{tabularx}
\usepackage{geometry}
\usepackage{subcaption}
\usepackage{footnote}
\usepackage{footmisc}
\makesavenoteenv{tabular}

\begin{document}
\thispagestyle{empty}
\begin{center}
\begin{minipage}{0.75\linewidth}
    \centering
    \includegraphics[width=0.6\linewidth]{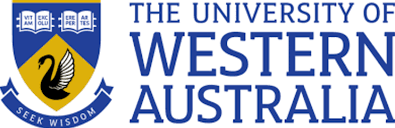}
    \par
    \vspace{2cm}
    {\uppercase{\large {\bf Analysis and Evaluation \\of Kinect-based \\Action Recognition Algorithms\par}}}
    \vspace{2cm}

    {\large MPE Research Thesis \\
    submitted in partial fulfilment of the requirements \\
    for the Degree of Master of Professional Engineering\par}
    \vspace{2cm}
    {\large presented to\\
    School of Computer Science and Software Engineering\\
    Faculty of Engineering, Computing Mathematics\\
    The University of Western Australia \par}
    \vspace{2cm}
    
    {\large Supervisor: A/Prof. Du Q. Huynh\par}
    \vspace{1cm}
    
    {\large by\\
    Lei Wang\par}
    \vspace{2cm}
    
    {\large October 2017}
\end{minipage}
\end{center}
\clearpage

\pagebreak

This MPE Research Thesis represents my own work and due acknowledgment is given whenever information is derived from other sources. No part of this MPE Research Thesis has been or is being concurrently submitted for any other qualification at any other university.

\vspace{3cm}

Signed:

\hspace{5em} 
\includegraphics[scale=0.4]{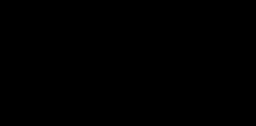} \\
\hspace{5em} 

\pagebreak

\chapter*{\centering Acknowledgment}

I am grateful to Associate Professor Du Huynh for her valuable suggestions and discussions. I also thank the authors of HON4D~\cite{Oreifej2013}, HOPC~\cite{Rahmani2016} and RBD~\cite{Vemulapalli2016} for making their techniques publicly available.

\addcontentsline{toc}{chapter}{Acknowledgment}

\pagebreak

\chapter*{\centering Abstract}

Human action recognition from videos is a challenging problem as the videos may be taken from different viewpoints and under different lighting conditions. Furthermore, the human subject carrying out the actions may have different body sizes and the speed of executing the actions may differ among individuals. To tackle these challenges, the Kinect depth sensor has been developed to record real time depth sequences, which are insensitive to the colour of human clothes and illumination conditions. Many methods on recognizing human action have been reported in the literature such as HON4D~\cite{Oreifej2013}, HOPC~\cite{Rahmani2016}, RBD~\cite{Vemulapalli2016} and HDG~\cite{Rahmani2014}. These methods use the 4D surface normals, pointclouds, skeleton-based model and depth gradients, respectively, to capture discriminative information from depth videos or skeleton data. In this research project, the performance of four aforementioned algorithms will be analyzed and evaluated using five benchmark datasets, which cover noise and the above challenging issues. We also implemented and improved the HDG algorithm, and applied it in cross-view action recognition using the UWA3D Multiview Activity dataset. Moreover, we used different combinations of individual feature vectors in HDG for performance evaluation. The experimental results show that our improvement of HDG outperforms other three state-of-the-art algorithms for cross-view action recognition.

\addcontentsline{toc}{chapter}{Abstract}

{\bf Keywords:} Evaluation, Kinect-based algorithms, cross-view action recognition.

\pagebreak

\tableofcontents

\pagebreak

\chapter{Introduction}

Human action recognition has been widely applied in human computer interaction, smart video surveillance, sports and health care; however, there still exist many challenging problems. Different viewpoints, visual appearances, human body sizes, lighting conditions, and speeds of action execution can have significant effects on the performance of classification and recognition methods. Furthermore, the human subject can be partially occluded by other objects in the scene. In order to tackle the challenges, the Kinect depth sensor has been developed to record real time depth sequences, which allow the foreground human subject to be better segmented than if conventional colour cameras are used. Depth images are also more advantageous for action recognition since they provide extra body shape information, which includes more discriminative information for detection, classification and recognition. In particular, depth images captured by the Kinect sensor are insensitive to illumination conditions and the colour of human clothes. Although many methods on recognizing human action from depth videos have been reported in the literature~\cite{Li2010, Shotton2011, Yang2012}, the performance of these methods is still greatly affected by issues such as loose clothing worn by the human subjects, different performing styles and speeds of the same action by different people, occlusion by objects in the scenes, and self-occlusion.

It is significant to design an effective algorithm to extract and represent corresponding characteristics from 3D video sequences for computer vision challenges. In particular, robust action recognition algorithms can be applied to a variety of recognition problems. For instance, \citet{Oreifej2013}~captured the discriminative features by projecting the 4D surface normals from the depth sequence onto a 4D regular space to build the histogram of oriented 4D normals (HON4D) descriptor. \citet{Rahmani2014}~put forward an algorithm which uses the histogram of depth gradients (HDG) that combines discriminative information from both depth images and 3D joint positions to deal with local occlusions so as to increase the accuracy of recognition. In their later research paper~\cite{Rahmani2016}, to deal with the sensitivity of viewpoint variations in recognition, they directly process pointclouds and compute the histogram of oriented principle components (HOPC) descriptor for view invariance action recognition. These two proposed algorithms have been shown to be more robust, efficient and accurate than several other algorithms at that time. At about the same year, \citet{Vemulapalli2014}~proposed a skeleton representation involving translations and rotations to model the relative 3D geometric relationships between body parts. Based on this work, they proposed a skeletal representation which models human actions as curves in the Lie Group $SO_3 \times ... \times SO_3$ ~\cite{Vemulapalli2016}. From here on, we refer to this representation as the rotation-based descriptor (RBD). These two methods~\cite{Vemulapalli2014, Vemulapalli2016} use the skeleton model only and they also perform well in some benchmark datasets such as the 3D Action Pairs dataset~\cite{Oreifej2013}.

In this research project, the performance of action recognition using the Kinect depth sensor for the four aforementioned algorithms~\cite{Oreifej2013, Rahmani2014, Vemulapalli2016, Rahmani2016}, which use the 4D surface normals, depth gradients, pointclouds and skeleton model respectively, will be evaluated using 5 benchmark human activity datasets. These selected activity datasets contain noise, different views, easily confused action classes, different action speed, different body size, background clutters and occlusion issues. The performance of these four algorithms will be evaluated and compared based on the recognition accuracy. 

\chapter{Related Work}

Action recognition methods can be classified into three categories based on the type of input data: colour-based~\cite{Li2010, Sung2011, Sung2012, Koppula2013, Rahmani2015}, depth-based~\cite{Yang2012, Oreifej2013, Rahmani2014, RahmaniHOPC2014, RahmaniLLC2014, RahmaniPRL2016, Rahmani2016}, and skeleton-based~\cite{Shotton2011, Sung2012, Xia2012, Vemulapalli2014, Vemulapalli2016, Amor2016, Rahmani2016CVPR}.  Early researches on action recognition from depth sequences attempted to use algorithms that are designed for colour sequences, when RGB-D cameras first became available. For example, \citet{Li2010} adopted an {\it action graph} to capture the human motion from a set of shared prominent postures among different actions. They mapped the depth image onto 3 orthogonal Cartesian planes which are $xy$, $zx$, and $zy$, and sampled a specified number of 2D points as a bag of representative 3D points at equal distance along the contours of the projections. Compared to other 2D silhouette recognition methods at that time, their approach gave better accuracy of recognition and was more robust to occlusion. However, their method required a large training set of all classes and was computationally expensive. 

Existing skeleton-based action recognition can be grouped into two categories: body part based methods and joint based methods. Joint based methods model the motion of joints (either individual or a combination) using features extracted by the OpenNI tracking framework, such as relative joint positions~\cite{Yang2012, Vemulapalli2014}, joint orientations with a fixed coordinate axis~\cite{Xia2012}, etc. Body part based methods represent the human body as an articulated system where rigid body parts are connected by joints. The connected rigid parts are represented as features such as joint angles~\cite{Sung2012}, temporal evolution of body parts~\cite{Amor2016, Rahmani2016CVPR}, 3D relative geometric relationships between rigid body parts~\cite{Vemulapalli2014, Vemulapalli2016}, etc. For example, \citet{Vemulapalli2016} proposed a new skeleton descriptor which represents the relative 3D geometric relationships using translations and rotations between rigid body parts in the 3D space. They modelled human actions as a curved manifold in a Lie group. After that, they unwrapped the action curves onto a vector space, which combines the logarithm map with rolling maps, and performed the action classification in Lie algebra. However, since skeleton-based methods are sensitive to action speed, like other skeleton-based methods, the performance of their method was affected when the recognition was performed on videos captured from different viewpoints.

Depth-based action recognition algorithms in the literature can be divided into two types: one category focuses on researching techniques based on depth images~\cite{Oreifej2013, Rahmani2014} and another focuses on the joint positions~\cite{Li2010, Yang2012, Rahmani2014}. Making full use of depth images in action recognition has become a popular research area, especially when the Microsoft Kinect sensor was released with the associated SDK. Three dimensional joint positions from depth images can be estimated with improved accuracy; correspondingly, the number of skeleton based recognition papers~\cite{Shotton2011, Yang2012, Rahmani2014} has increased. \citet{Shotton2011} proposed a new technique, which treats pose estimation as object recognition based on intermediate body parts from single depth image without using temporal information. They turned the pose estimation problem into a simpler classification problem, and the large varied training data makes the estimation classifier more robust to body shape, pose, etc. In addition, \citet{Yang2012} adopted the {\it EigenJoints} method which combines action information (such as posture, motion) and offsets features based on the differences of pairwise 3D joint positions. Their method was based on applying Principal Component Analysis (PCA) on joint differences, and they tackled the multi-class actions problem by adopting a Naive Bayes Nearest Neighbour (NBNN) classifier. However, their recognition accuracy was compromised because the 3D joints alone could not capture enough discriminative information. 

Depth-based human action recognition techniques can also be classified into holistic~\cite{Li2010, Oreifej2013} and local methods~\cite{RahmaniHOPC2014, Rahmani2016} depending on whether the features used in the classification process are global or local. Holistic approaches for depth images are popular recently and the global feature descriptor is obtained from the entire depth sequence, which computes the motion and shape features together for action recognition. For instance, \citet{Oreifej2013} presented a global feature descriptor to capture the geometry and motion in the 4D space. They used the vertices of a polychoron as projectors to compute the distribution of normals for each sequence. Although the experiments indicated their descriptor outperform some relevant benchmarks, strong derivative information from silhouettes and edges might be suppressed. On the other hand, \citet{RahmaniHOPC2014}~presented a more informative descriptor which captures the data in three principal directions. Their method processed the pointclouds directly to extract HOPC descriptor from the local geometric features around each point within both the spatial support volume and spatio-temporal support volume. Furthermore, they proposed an approach to capture spatio-temporal keypoints (STK)~\cite{Rahmani2016} in the sequences and used local HOPC descriptors for action recognition at key locations only, which was more robust to noise, viewpoint variations as well as action speed. They demonstrated in their experimental results that their algorithm had significant improvements, especially for cross-view action recognition.

Other approaches such as~\cite{RahmaniLLC2014, RahmaniPRL2016} solved the action recognition problem for both depth and colour videos by capturing enough discriminative characteristics. For a conventional video sequence, they divided the sequence into hierarchical structures, which consist of a set of equal-sized subsequences. Each subsequence was composed of equal-sized blocks, and each block was composed of a number of equal-sized cells. The histogram of oriented 3D gradients (HOG3D) was computed in each cell, and then all cell descriptors in each block was concatenated to a block descriptor for that block. The locality-constrained linear coding (LLC) was applied efficiently in each block to reduce the intra class variations. Their experimental results showed consistently high accuracy for depth and conventional videos comparing with ten state-of-art techniques. \citet{Rahmani2014}~also presented another algorithm which combines depth and depth gradient features with joint position difference features and joint movement volume features for action classification. They applied a Random Decision Forests (RDF) to prune and classify all the features into corresponding classes. In general, their proposed algorithm achieved better results than other state-of-art algorithms and achieved a processing speed of more than 112 frames per second.

Motivated by these algorithms which focus on different aspects of features recorded by the Kinect camera for action recognition, the performance of four algorithms: the HON4D method proposed in~\cite{Oreifej2013}, the HDG method proposed in~\cite{Rahmani2014}, the HOPC method presented in~\cite{Rahmani2016}, and the RBD method proposed in~\cite{Vemulapalli2016} will be evaluated and compared on 5 standard action datasets. 

\chapter{Algorithms to be Analyzed and Evaluated}

Four algorithms will be evaluated in this project. The HON4D, HOPC and RBD codes were obtained from the original authors' website. However, the HDG code was implemented in Matlab in this research project since it is not publicly available.

\section{The HON4D algorithm}

\begin{figure}[H]
\begin{center}
\includegraphics[width=0.9\textwidth]{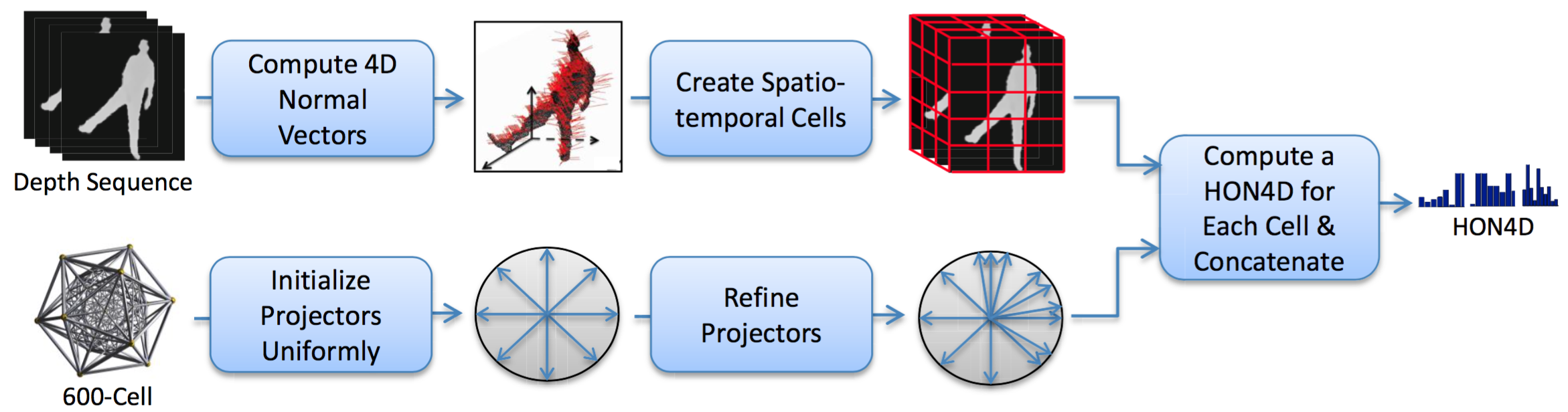} \caption{Steps for the HON4D descriptor computation~\cite{Oreifej2013}.}
\label{hon4d}
\end{center}
\end{figure}

\citet{Oreifej2013} presented a global feature descriptor to capture the geometry and motion of human action using a histogram of oriented 4D normals in the space of spatial coordinates, depth and time (see figure~\ref{hon4d}). They quantised a 4D space using a 600-cell polychoron with 120 vertices, and these vertices were used as projectors to compute the distribution of normals for each sequence. In order to increase the difference between two similar action classes, more vertices were induced randomly as discriminative projectors according to the density of the distribution. 

\section{The HDG algorithm}

\begin{figure}[tbp!]
\begin{center}
\includegraphics[width=0.9\textwidth]{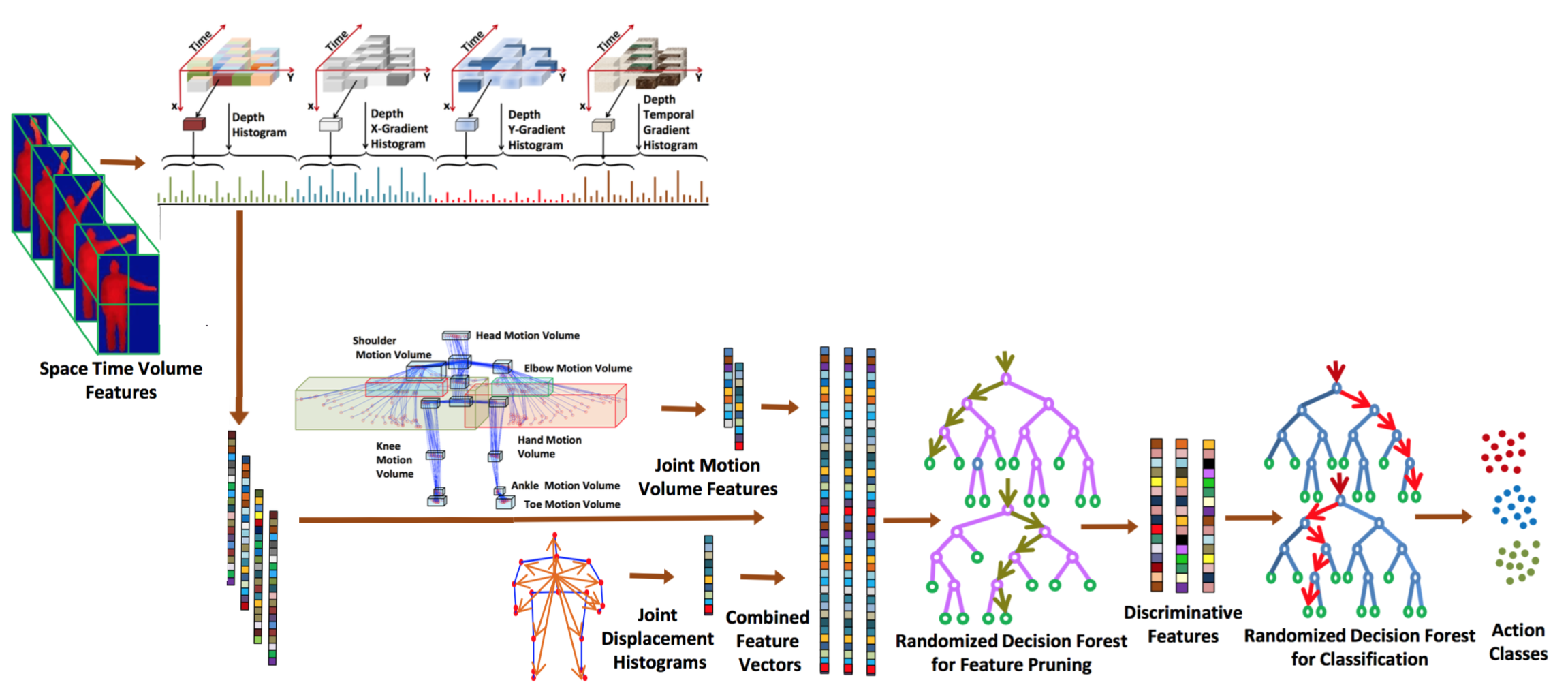} \caption{The HDG for real-time action recognition~\cite{Rahmani2014}.}
\label{hodrdf}
\end{center}
\end{figure}

\citet{Rahmani2014} proposed an algorithm for action recognition by concatenating local discriminative information from both depth images and 3D joint positions to increase recognition accuracy. Figure~\ref{hodrdf} shows the algorithm they proposed in their paper.

The final HDG feature vector was a combination of 4 components, and the main steps for computing each component are shown below:

\begin{itemize}
\setlength{\itemsep}{0ex}
\item Histogram of depth (hod):
\begin{itemize}
\setlength{\itemsep}{0ex}
\item Dividing each video sequence into a fixed number of subvolumes;
\item Calculating the depth range using the maximum and minimum depth values found in the video sequence;
\item Recording the frequency of depth range in each subvolume as a depth histogram vector;
\item Concatenating each depth histogram vector as a global depth feature vector for an action.
\end{itemize}
\item Histogram of depth derivatives (hodg):
\begin{itemize}
\setlength{\itemsep}{0ex}
\item Calculating the depth derivatives (along $x$-, $y$- and $t$- axes) for each frame in each input video sequence;
\item Calculating the derivative ranges and then dividing them into a uniform step size;
\item Recording the derivative ranges for each subvolume as depth derivative feature vectors;
\item Concatenating all depth derivative vectors into a global derivative feature vector.
\end{itemize}
\item Histogram of joint position differences (jpd):
\begin{itemize}
\setlength{\itemsep}{0ex}
\item Using spatial coordinates, depth and time to define the 3D joint position;
\item Finding the distance of each joint from a reference joint (such as the torso joint) along spatial coordinates and depth direction in each frame;
\item Making one histogram for each component (spatial coordinates and depth) for a video sequence;
\item Concatenating these 3 components to make a global skeleton feature vector.
\end{itemize}
\item Histogram of joint movement volumes (jmv):
\begin{itemize}
\setlength{\itemsep}{0ex}
\item Computing a 3D joint volume occupied by each joint;
\item Finding extreme positions along spatial coordinates and depth direction in the whole video sequence by calculating maximum position differences;
\item Incorporating the joint volume and extreme positions in the feature vector for each joint;
\item Concatenating all joint feature vectors to form a global joint movement feature vector.
\end{itemize}
\end{itemize}

Two RDFs were trained in this algorithm, one was used for feature dimension reduction and the other one was used for classification.

\section{The HOPC algorithm}

\citet{Rahmani2016} directly processed the 3D pointclouds to get the HOPC descriptor with STKs in the sequences, which is robust to viewpoint, scale variations, noise, etc. For a sequence of 3D pointclouds, the HOPC descriptor was extracted at each point in that sequence. For each point in the sequence, they defined two types of support volume: the {\it spatial support volume} and the {\it spatio-temporal support volume}. They then applied the PCA to all the points within a support volume to get the matrix of eigenvectors and corresponding eigenvalues. The ratio between the two largest eigenvalues was used to determine whether a STK will form or not. Rahmani et al. computed local HOPC at STKs only so as to make the descriptor more robust to viewpoints. They projected each eigenvector onto 20 projectors, which are 20 vertices of a regular dodecahedron. After applying a cut-off threshold for each bin and scaled by the corresponding eigenvalues, the HOPC descriptor was a combination of 3 small histograms for a specific point. They also defined a quality factor to detect significant motion variations in the spatio-temporal support volume. This process ensures that the same point in different views will have the same standard representation. 

\section{The RBD algorithm}

\begin{figure}[tbp!]
\begin{center}
\includegraphics[width=0.9\textwidth]{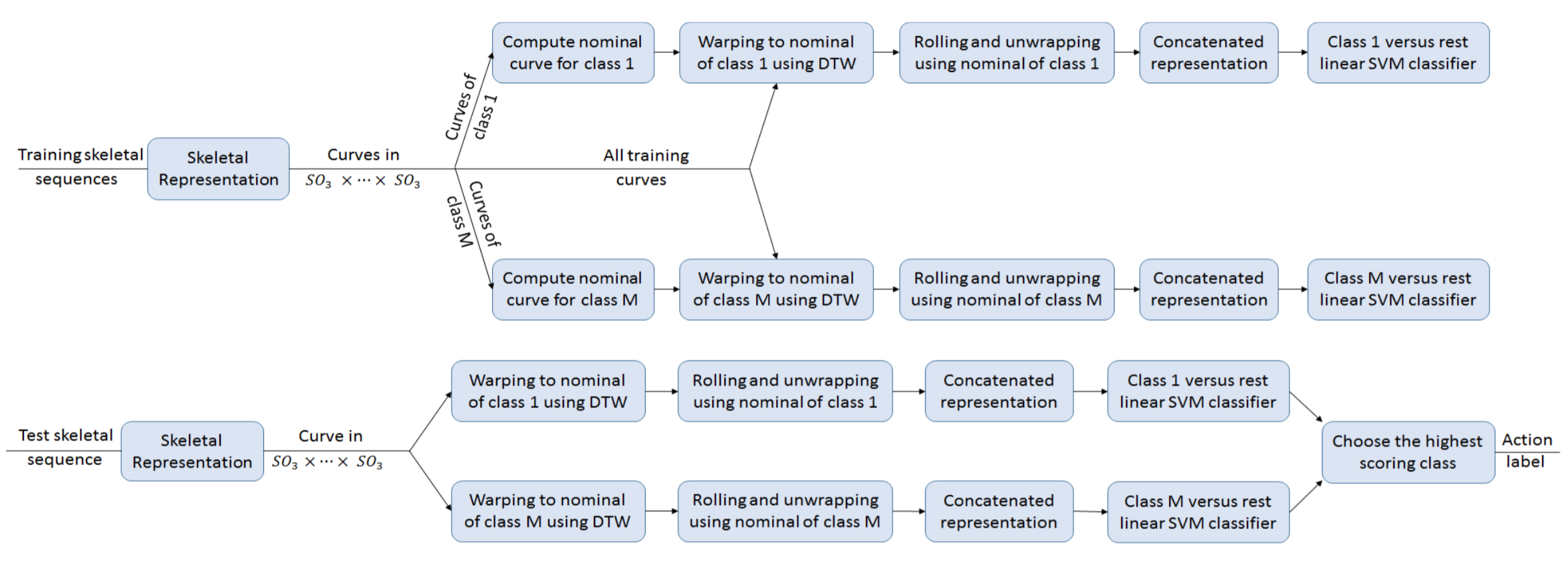} \caption{The training and testing phase using the RBD algorithm~\cite{Vemulapalli2016}.}
\label{pipeline}
\end{center}
\end{figure}

Figure~\ref{pipeline} shows the entire pipeline for computing the RBD as proposed in \cite{Vemulapalli2016}. The following 4 stages explain the algorithm in more details:
\begin{itemize}
\setlength{\itemsep}{0ex}
\item Skeleton representation: the relative 3D rotation between each pair of body parts was used to represent the 3D human skeleton. Since 3D rotations form the Lie group $SO_3$, the skeleton representation was then a point in these space of the direct product of Lie groups.
\item Nominal curves representation: after the skeleton representation was defined, human actions can be modelled as curves in the space described above. For each action class during the training stage, the authors computed a nominal curve and used the dynamic time warping (DTW) to wrap all curves;
\item Unwrapping and rolling stage: rolling each special orthogonal group $SO_3$ over its Lie algebra separately, and unwrapping the action curves over the Lie algebra;
\item Classification: concatenating all unwrapped action curves into a global feature descriptor and then classifying the features using a one-vs-all classifier.
\end{itemize}

\chapter{Experimental Settings}

\section{Benchmark Datasets}

Five benchmark datasets (Table \ref{datasets}) have been used in this project, each of which is detailed below.

\begin{table}[H]
\caption{Detailed information for the 5 benchmark datasets.}
\begin{center}
\resizebox{0.9\textwidth}{!}{\begin{tabular}{| l | c | c | c | c | l | c |}
\hline
 Datasets & Classes & Subjects & Views & Sensor & Modalities & Year\\ 
\hline
\hline
MSRAction3D Dataset & 20 & 10 & 1 & Kinect v1 & Depth + 3DJoints & 2010 \\
\hline
3D Action Pairs Dataset & 12 & 10 & 1 & Kinect v1 & RGB + Depth + 3DJoints & 2013\\
\hline
Cornel Activity Dataset (CAD-60) & 14 & 4 & - & Kinect v1 & RGB + Depth + 3DJoints & 2011 \\
\hline
UWA3D Single View Dataset & 30 & 10 & 1 & Kinect v1 & RGB + Depth + 3DJoints & 2014 \\
\hline
UWA3D Multiview Dataset & 30 & 9 & 4 & Kinect v1 & RGB + Depth + 3DJoints & 2015 \\
\hline
\end{tabular}}
\label{datasets}
\end{center}
\end{table}

\begin{enumerate}
\setlength{\itemsep}{0ex}
\item The {\bf MSRAction3D Dataset}~\cite{Oreifej2013} is an action dataset captured by the Kinect depth camera. It contains 20 human sports activities such as {\it jogging}, {\it golf swing} and {\it sideboxing}. Each action in this dataset is performed by 10 people for 2 or 3 times. This dataset also includes the skeleton data~\cite{Vemulapalli2014}.
\item The {\bf 3D Action Pairs Dataset}~\cite{Oreifej2013} contains actions selected in pairs, each pair of two actions have similar motion trajectories and shape such as {\it put on a hat} and {\it take off a hat}. This dataset also has the skeleton data~\cite{Vemulapalli2016}.
\item The {\bf Cornell Activity Dataset} (CAD)~\cite{Koppula2013, Sung2012, Sung2011}  comprises two sub-datasets, CAD-60 and CAD-120. Both sub-datasets contain RGB-D and tracked skeleton video sequences of human activities captured by a Kinect sensor. In this research project, only CAD-60 is used in the experiments. Figure~\ref{cad60examples} shows some depth images from CAD-60.

\begin{figure}[tbp!]
    \centering 
\begin{subfigure}{0.3\textwidth}
  \includegraphics[width=\linewidth]{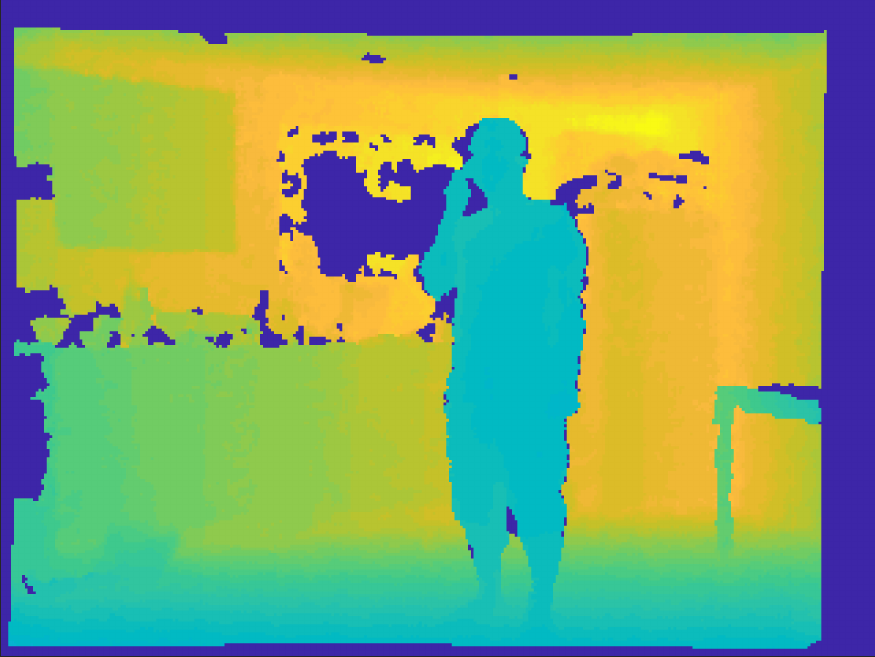}
  \caption{talking(phone)}
  \label{fig:1}
\end{subfigure}\hfil 
\begin{subfigure}{0.3\textwidth}
  \includegraphics[width=\linewidth]{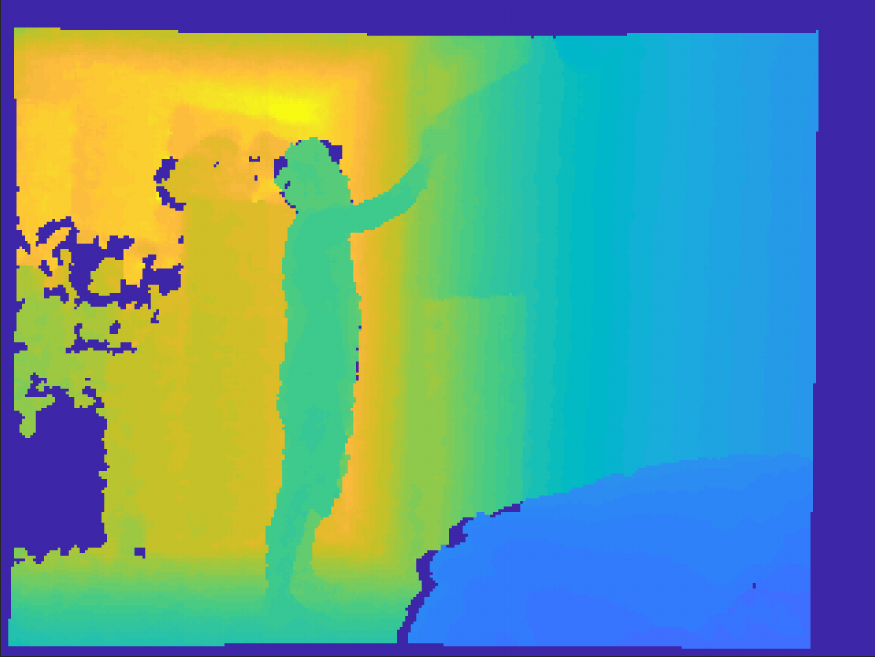}
  \caption{writing}
  \label{fig:2}
\end{subfigure}\hfil 
\begin{subfigure}{0.3\textwidth}
  \includegraphics[width=\linewidth]{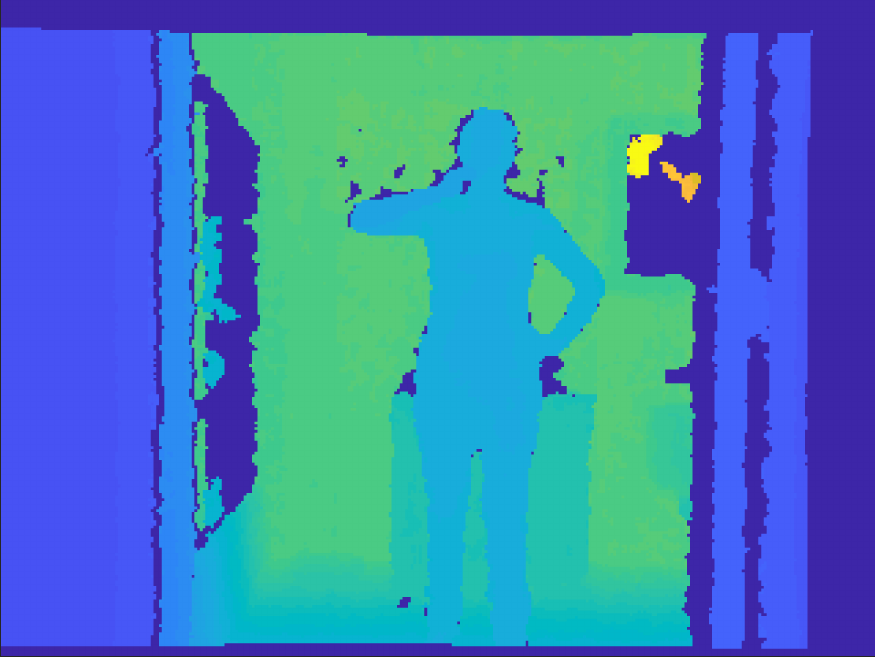}
  \caption{brushing teeth}
  \label{fig:3}
\end{subfigure}
\medskip
\begin{subfigure}{0.3\textwidth}
  \includegraphics[width=\linewidth]{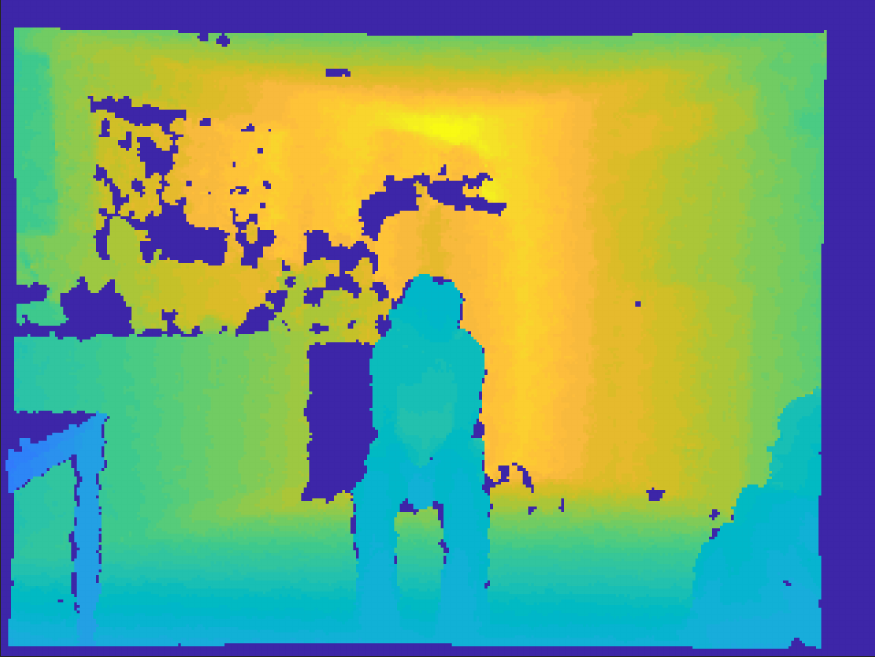}
  \caption{talking(couch)}
  \label{fig:4}
\end{subfigure}\hfil
\begin{subfigure}{0.3\textwidth}
  \includegraphics[width=\linewidth]{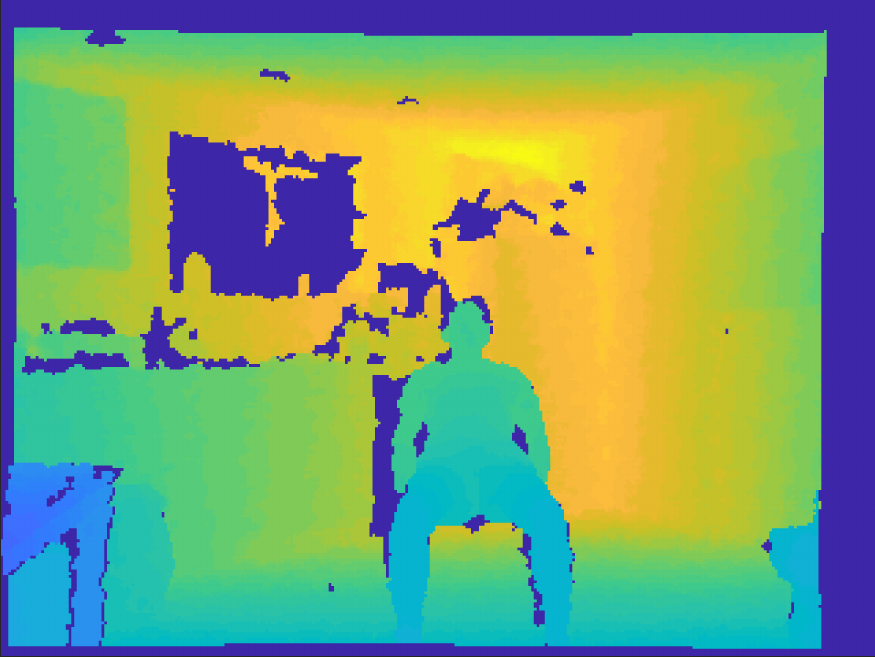}
  \caption{relaxing(couch)}
  \label{fig:5}
\end{subfigure}\hfil 
\begin{subfigure}{0.3\textwidth}
  \includegraphics[width=\linewidth]{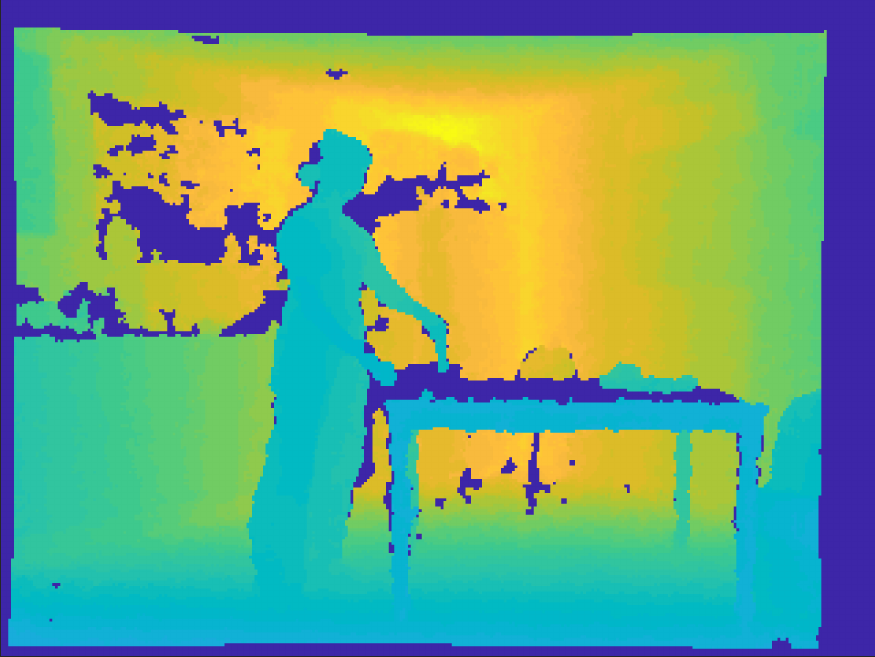}
  \caption{cooking(stirring)}
  \label{fig:6}
\end{subfigure}
\caption{Sample depth images from CAD-60.}
\label{cad60examples}
\end{figure}

\item The {\bf UWA3D Single View Dataset}~\cite{Rahmani2014} contains 30 actions performed by 10 people in cluttered scenes. This dataset includes the skeleton data and depth sequences.
\item The {\bf UWA3D Multiview Activity Dataset}~\cite{Rahmani2016} contains 30 actions performed by 9 people with 4 different viewpoints in cluttered scenes. This dataset also has the skeleton data and depth sequences.
\end{enumerate}

The first 4 datasets have been used for single view action recognition. Since the UWA3D Multiview Dataset has 4 views, a cross-view recognition strategy has been applied on this dataset. 

\section{Evaluation Settings}

All 4 algorithms~\cite{Oreifej2013, Rahmani2014, Vemulapalli2016, Rahmani2016} have been tested in Matlab on the Windows platform. Table~\ref{dataformat} shows the data type used in each algorithm.

\begin{table}[tbp!]
\caption{Data type used in each algorithm}
\centering
\resizebox{0.6\textwidth}{!}{\begin{tabular}{| c | c |}
\hline
HON4D\cite{Oreifej2013} & depth sequence/avi files\\
\hline
HDG\cite{Rahmani2014} & depth sequence and skeleton data(txt files)\\
\hline
HOPC\cite{Rahmani2016} & .mat files (Matlab file for depth sequence)\\
\hline
RBD\cite{Vemulapalli2016} & training/testing sets \& skeleton data (.mat)\\ 
\hline
\end{tabular}}
\label{dataformat}
\end{table}

\begin{itemize}
\item For the HON4D algorithm, there are 3 types of descriptors which are included in the code directory: HON4D which uses the projectors from a polychoron as described in \cite{Oreifej2013}, HON4DA which uses 3 angles to quantize the surface 4D normals but not described in~\cite{Oreifej2013} and LocalHON4D which is similar to HON4D. The first two (HON4D and HON4DA) are C++ codes and need OpenCV. In~\cite{Oreifej2013}, the frame size for all datasets in the experiments was $320 \times 240$ and each video was divided into $4 \times 3 \times 3$ spatiotemporal cells (${\text {width}} \times {\text {height}} \times {\text {\#frames}}$). In this project, the same settings have been used for evaluation.

\item For the HOPC algorithm, there are 3 separate functions: pointclouds extraction, feature extraction and classification. In feature extraction, \citet{Rahmani2016} used different spatial scale for different datasets. For example, spatial scale for MSRAction 3D was 19, but for UWA3D dataset the spatial scale was changed to 140 in their experiments. Moreover, they divided each depth video into $6 \times 5 \times 3$ (along $X$, $Y$ and $T$) spatiotemporal cells to extract features. In this project, constant temporal scale and spatial scale have been used for 5 benchmark datasets. For MSRAction3D and 3D Action Pairs datasets, temporal scale and spatial scale were set as 2 and 19 respectively; for other datasets, we used 2 as temporal scale and 140 as spatial scale. 

\item For the RBD algorithm, they used half of the subjects for training and the remaining half for testing. There is a function containing the entire pipeline for action recognition: skeleton representation, temporal modelling (DTW, mapping to Lie algebra and Fourier temporal pyramid representation (FTP)) and one-vs-all classification. Some datasets should be prepared such as body model, labels of training and testing sets, and the whole skeleton data should be saved as .mat files before running the algorithm. In addition, the desired frames for skeleton representation should be changed based on different datasets. For example, the desired frame for 3D Action Pairs datasets was 111 in~\cite{Vemulapalli2016}. In this project, the desired frame for both UWA3D Single View and Multiview Activity datasets was set as 100; the desired frames for MSRAction3D and 3D Action Pairs datasets were set as 76 and 111 respectively based on~\cite{Vemulapalli2014, Vemulapalli2016}; and the desired frame for CAD-60 was 1000.

\item For the HDG algorithm, based on the research given in~\cite{Rahmani2014}, the number of subvolumes does not contribute to any discriminative features. Therefore, in this project, each video sequence in each dataset has been divided into $10 \times 10 \times 5$ (along $X$, $Y$ and $T$) subvolumes for computing the depth and depth gradient histograms. For the joint movement volume features, each joint volume has been divided into  $1 \times 1 \times 5$ (along $X$, $Y$ and $T$) cells. There are 4 individual feature vectors in the HDG algorithm, which are: 
\begin{itemize}
\item histogram of depth (hod), 
\item histogram of depth gradients (hodg), 
\item joint position differences (jpd), 
\item joint movement volume features (jmv).
\end{itemize}
The performance of different combinations of individual feature vectors have been evaluated on 5 benchmark datasets, and a cross-view action recognition strategy will be applied on the UWA3D Multiview Activity dataset.
\end{itemize}

Since the code for the HDG algorithm is not available, a part of this project is to improve and implement this algorithm in Matlab.

\section{Evaluation Measure}

Average recognition accuracy for a given class $A$ is given by
\begin{equation}
\text{average recognition accuracy}=\frac{\text{\# correct class $A$ labels from algorithm}}{\text{\# total class $A$ labels in dataset}}.
\label{eq:accuracy}
\end{equation}
The measure given in Equation~\ref{eq:accuracy} is for one class only. To illustrate the recognition accuracy of an algorithm, a confusion matrix shown the recognition accuracy of all classes is often used.

\chapter{Optimisation of Hyperparameters for HDG}

There are 3 hyperparameters in HDG. One hyperparameter which is the number of subvolumes, the same value given in ~\cite{Rahmani2014} has been used. Other 2 hyperparameters which are the number of trees trained and threshold factor for feature pruning have been optimised in this project.

For the HDG algorithm, the total length of each global feature vector for the MSRAction3D dataset is 13250 and the length for hod, hodg, jmv and jpd are 2500, 10000, 600 and 150 respectively. Therefore, one RDF is trained to select predictors with higher importance values for feature dimension reduction. We do not use PCA for dimension reduction since this technique is sensitive to the scale of measurement, and RDF can deal with the missing skeleton data and identify significant predictors for the later classification. In this case, we can call MATLAB function {\it fitensemble} and {\it predictorImportance} to get each importance value for each predictor. After that, each feature importance value $p_i$ is normalized to $\hat{p_i}$ in the global feature vector using the equation:

\begin{equation}
\hat{p_i}= \frac {{p}_i}{{\lvert \lvert {p}_i\rvert \rvert}_2}, ~\text{for}~1 \leq i \leq~N, 
\label{eq:imp}
\end{equation}
where $N$ is the total number of predictors. The total number of predictors in the global feature vector depends on the datasets since different datasets have different number of body joints.
\begin{figure}[tbp!]
\begin{center}
\includegraphics[width=0.98\textwidth]{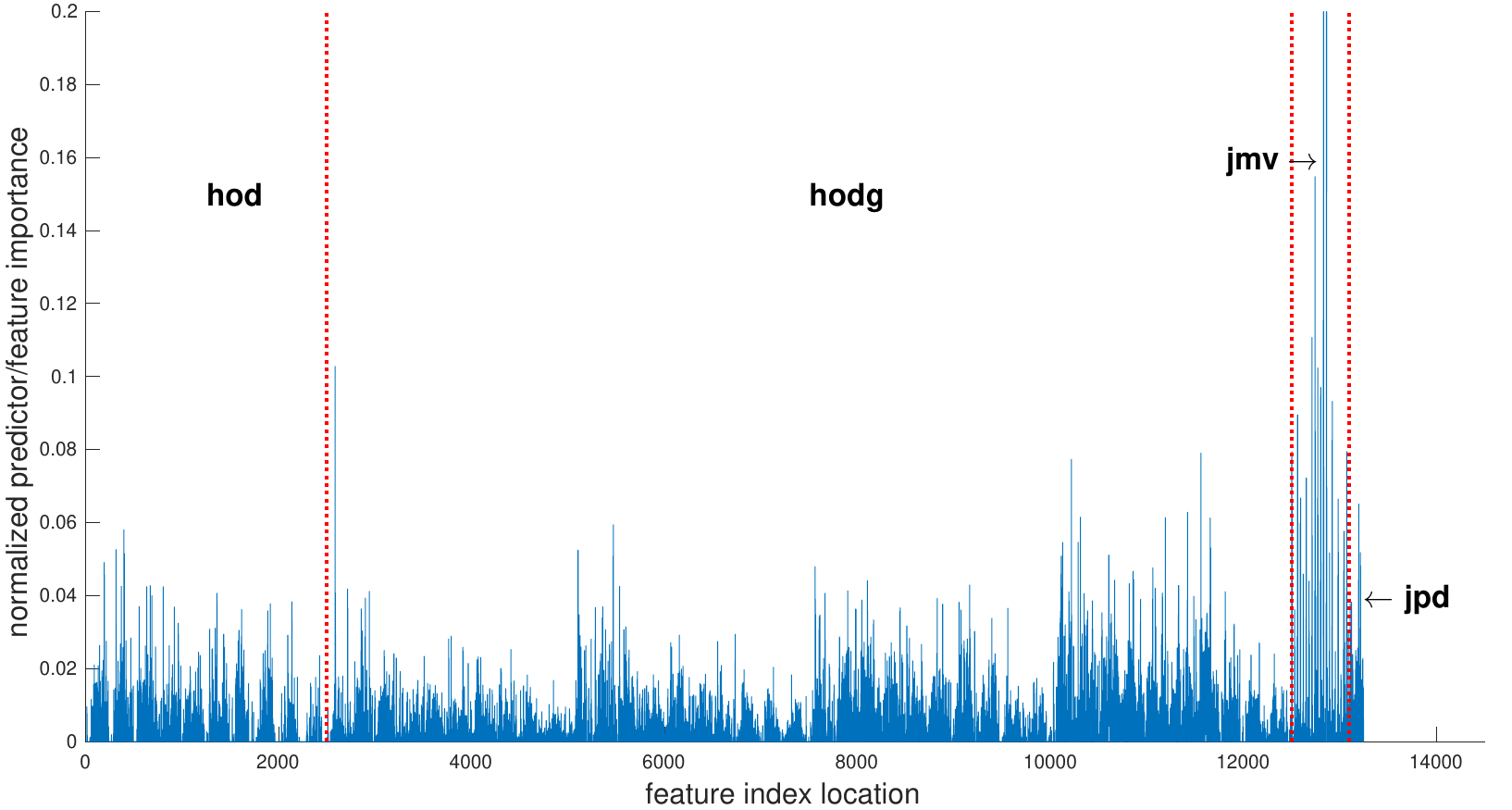}
\caption{Normalized feature importance of 13250 predictors with the original index preserved for the MSRAction3D dataset.}
\label{predictorimp}
\end{center}
\end{figure}

Figure~\ref{predictorimp} details the normalized feature importance with the original index preserved for the MSRAction3D dataset. As shown in the figure, some predictors have higher importance values. In each individual feature vector, these predictors capture more discriminative information. In addition, the jmv feature seems to have higher importance value comparing with other individual feature vectors. To reduce the length of global feature vector and retain key discriminant predictors, a threshold value $\theta$ for feature selection is given by:

\begin{equation}
\theta=\alpha \times \frac{\sum_{k=1}^\text{N} \hat{p_i}}{\text{N}},
\label{eq:thresholdvalue}
\end{equation}
where $\alpha$ is the threshold factor. Comparing each normalized feature importance value with this threshold value $\theta$ and if its importance value is greater than $\theta$, this predictor is kept for classification; otherwise it is discarded.

\begin{figure}[tbp!]
    \centering
    \begin{subfigure}[t]{0.48\textwidth}
        \centering
        \includegraphics[width=\textwidth]{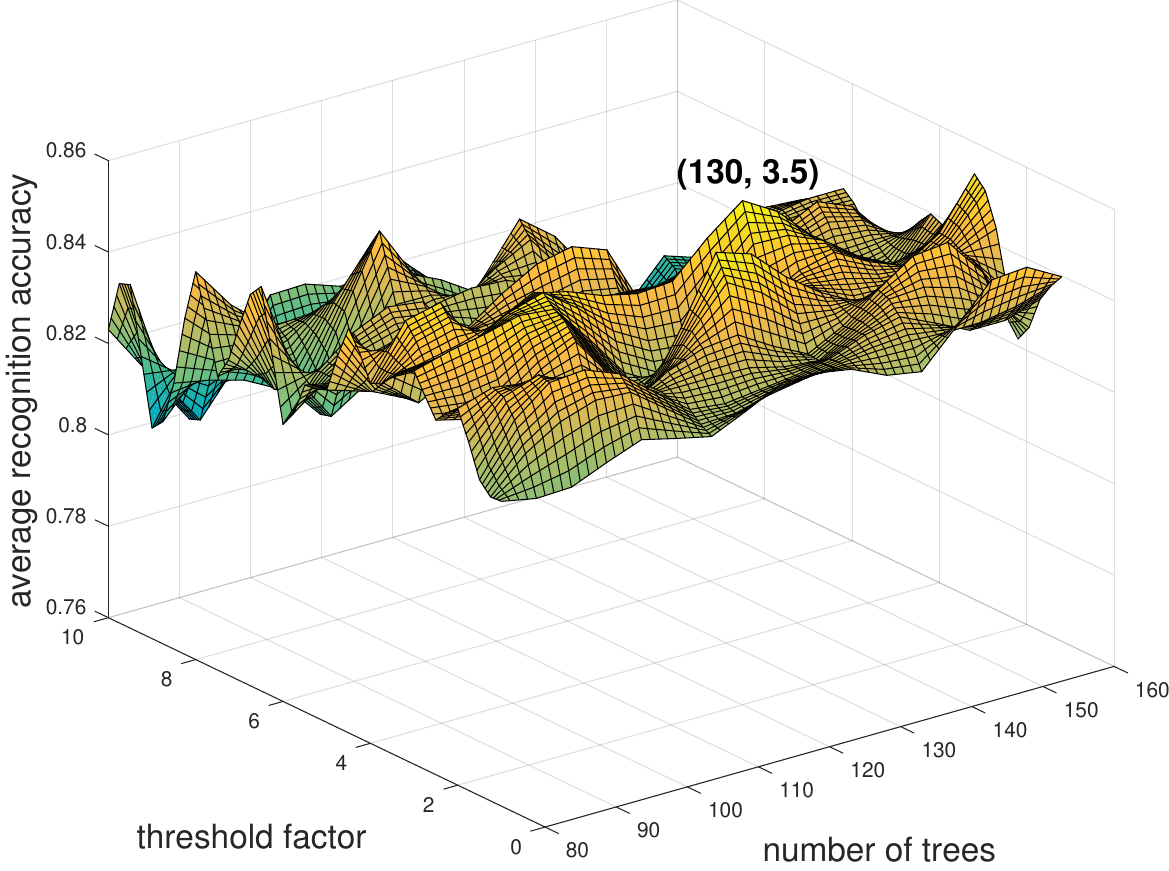}
        \caption{Surface plot.}
    \end{subfigure}
    \begin{subfigure}[t]{0.48\textwidth}
        \centering
        \includegraphics[width=\textwidth]{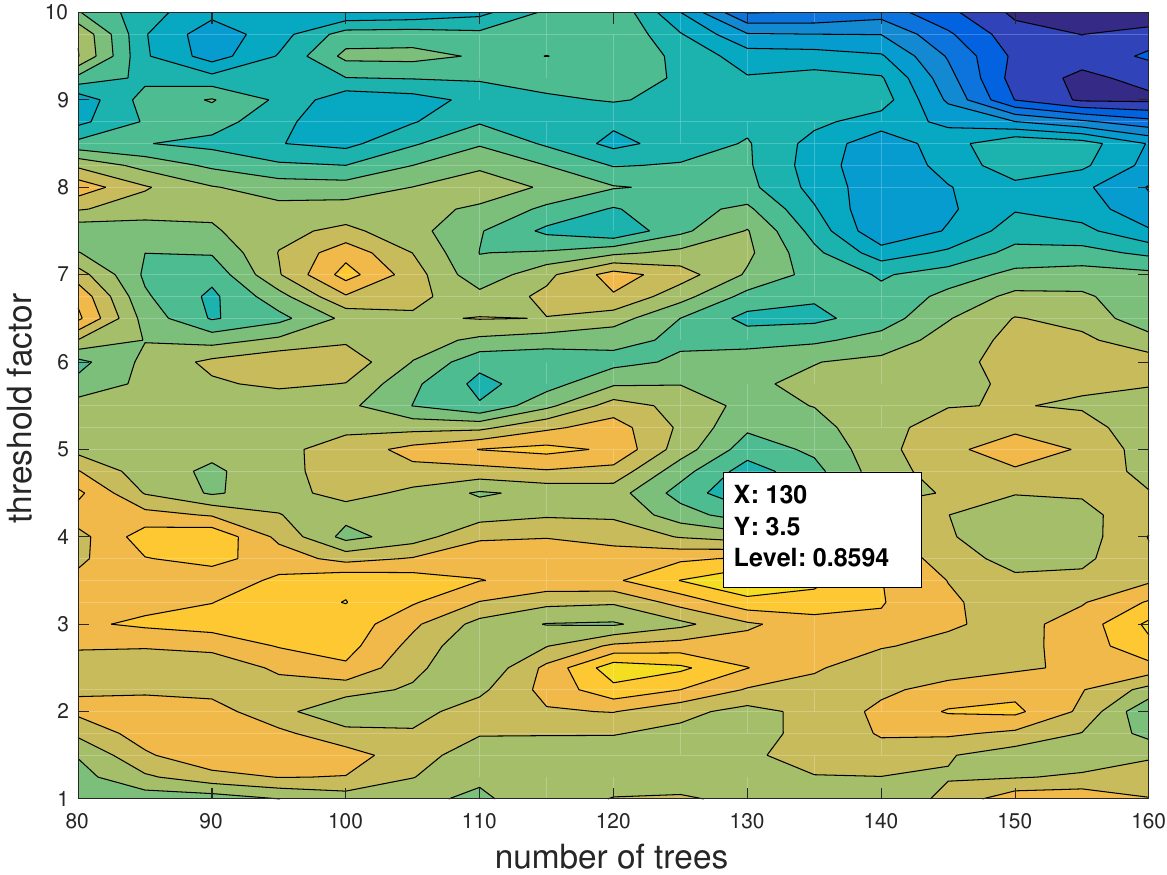}
        \caption{Contour plot.}
    \end{subfigure}
    \caption{Average recognition accuracy for the MSRAction3D dataset using HDG-all features with different number of trees trained and different threshold factors for feature selection.}
    \label{parameterperformance}
\end{figure}

After feature selection, the lower dimensional features are passed to the second RDF for classification. Based on the research published in~\cite{perner}, even if there is a larger number of trees (greater than 128) in a random forest, there is no desirable performance gain. Moreover, the experiments given in~\cite{perner} show that there is a good balance among performance, memory usage and handling time if the number of trees in a forest is between 64 and 128. Figure~\ref{parameterperformance} shows the relationship between the average recognition accuracy and two performance parameters (number of trees trained and threshold factor $\alpha$) for feature selection on the MSRAction3D dataset. It can be seen in the figure that the optimal number of trees trained for feature pruning is 130 and the optimal $\alpha$ is approximately 3.5 when all feature vectors are incorporated in HDG. 

To evaluate the performance of different combinations of individual feature vectors for the HDG algorithm, we find the optimal number of trees and the optimal threshold factor $\alpha$ for each different combinations of feature vectors based on the analysis of the MSRAction3D dataset, and use these optimal values for other datasets. For the UWA3D Multiview Activity dataset, we use view 1 and view 2 for training and view 3 for testing to evaluate and find optimal hyperparameters for each different combinations of feature vectors, and use these optimal values for cross-view action recognition experiments.

\chapter{Results and Discussions}

\section{Results and Discussions for the first 4 datasets}

For the first 4 datasets (MSRAction3D, 3D Action Pairs, CAD60 and UWA3D Single View datasets), half of the subjects' data are used for training and the remaining half for testing. All possible 252 (${\binom nk}={\binom {10}{5}=252}$) combinations of selections (the combinations of selections for CAD60 are 6 since there are only 4 subjects in this dataset) are used to calculate the average recognition accuracy. However, for the RBD~\cite{Vemulapalli2016} algorithm, 10 different random combinations of training and testing subjects are used for MSRAction3D, 3D Action Pairs and UWA3D Single View datasets (the combinations of selections for CAD60 are only 6 since there are only 4 subjects in this dataset) to get the average recognition accuracy based on the evaluation settings given in \cite{Vemulapalli2016}.

\begin{table}[H]
\caption{Comparison of average action recognition accuracy (percentage) on 4 benchmark datasets with different combinations of feature vectors incorporated in HDG.}
\begin{center}
\resizebox{0.9\textwidth}{!}{\begin{tabular}{| l | c | c | c | c | c |}
\hline
 & Action3D & ActionPairs & CAD60 & UWA Singleview\\ 
\hline
\hline
HOPC~\cite{Rahmani2016}  & 85.49 & 92.44 & 47.55 & 60.58\\
\hline
RBD-logarithm map \cite{Vemulapalli2016}& 88.69 & 92.96 & 69.12 & 51.96 \\ 
RBD-unwrapping while rolling \cite{Vemulapalli2016}& 88.47 & 94.09 & 69.12 & 53.05\\
RBD-FTP representation \cite{Vemulapalli2016}& 89.40 & 94.67 & 76.96 & 50.41\\ 
\hline
\hline
HDG-hod & 66.22 & 81.20 & 26.47 & 44.35\\
\hline
HDG-hodg & 70.34 & 90.98 & 50.98 & 54.23 \\ 
\hline
HDG-jpd & 55.54 & 53.78 & 46.08 & 40.88\\
\hline
HDG-jmv& 62.40 & 84.87 & 41.18 & 51.02\\ 
\hline
HDG-hod+hodg & 71.81 & 90.96 & {\bf 51.96} & 55.17\\
\hline
HDG-jpd+jmv & 65.57 & 84.93 & 49.02 & 55.57 \\
\hline
HDG-hod+hodg+jpd & 72.06 & 90.72 & 51.47 & 56.41\\
\hline
HDG-hod+hodg+jmv & {\bf 75.41} & {\bf 92.27} & 49.51 & 58.80\\
\hline
HDG-hodg+jpd+jmv & 75.00 & {\bf 92.28} & {\bf 52.94} & {\bf 59.82}\\
\hline
HDG-all features & {\bf 75.45} & 92.13 & {\bf 51.96} & {\bf 60.33}\\
\hline
\end{tabular}}
\label{evaluationresults}
\end{center}
\end{table}

Table~\ref{evaluationresults} compares the implemented HDG algorithm with existing approaches, and two highest recognition accuracy for the HDG algorithm in each column have been highlighted. We observe from the table that HDG does not achieve higher recognition accuracy comparing with other competitors. There are several reasons. Firstly, the recognition accuracy for RBD~\cite{Vemulapalli2016} is only averaged over 10 different randomly combined training and testing subjects, and skeleton data with lower confidence values are discarded and not used during training and testing stages. Secondly, HOPC~\cite{Rahmani2016} is a robust descriptor since it is extracted directly from pointclouds, which is robust to noise and change of action speed. Thirdly, hod and hodg in HDG are extracted from preprocessed depth sequences which still contain some noise, although human subjects have been segmented. In addition, we do not discard skeleton data with lower confidence values and in contrast, we use them during training and testing stages. 

As we can see from the table, RBD~\cite{Vemulapalli2016} gives the best results on the MSRAction3D, 3D Action Pairs and CAD60 datasets, it outperforms HDG by 24.02\% and outperforms HOPC~\cite{Rahmani2016} by 29.41\% on the CAD60. For the UWA3D Single View dataset, HOPC~\cite{Rahmani2016} performs better than HDG-all features and RBD~\cite{Vemulapalli2016} since it encodes the motion speed in a robust way and this descriptor does not rely on the depth gradients, which is more robust to noise.

\begin{figure}[tbp!]
\begin{center}
\includegraphics[width=0.98\textwidth]{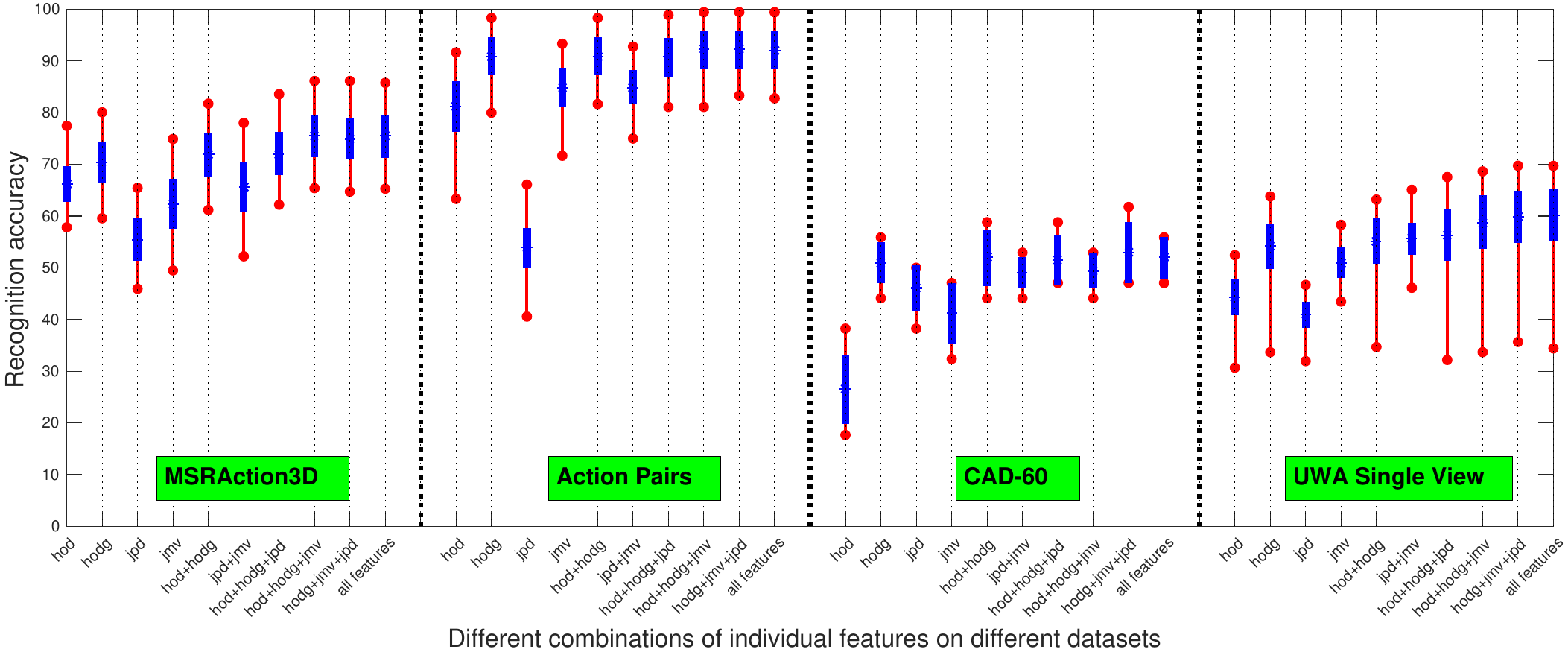}
\caption{Box plot for the HDG algorithm with different combinations of individual feature vectors used on 4 benchmark datasets.}
\label{boxplot}
\end{center}
\end{figure}

Figure~\ref{boxplot} shows a box plot for the maximum, minimum, standard deviation and mean values of recognition accuracy on 4 benchmark datasets with different evaluation settings for the HDG algorithm. We observe from the plot that concatenating more individual feature vectors contributes to higher recognition accuracy; however, incorporating a lower recognition accuracy component (individual feature vector) in the global feature vector for classification may decrease the performance of HDG. In general, the hodg feature vector captures more discriminative information than the hod feature vector as can be found in Table~\ref{evaluationresults}. Moreover, the jmv feature vector can not only capture the motion trajectories but also the sequence of actions, which is more robust than the jpd feature vector. Therefore, the jmv feature vector is of vital importance for contributing to higher recognition accuracy, which is consistent with the results of predictor importance shown in Figure~\ref{predictorimp}.

Since there are only 4 subjects in CAD60 dataset and only half subjects are used in the training stage, the total recognition accuracy is lower than other datasets. Moreover, the depth videos given in this dataset contains noise and background clutters (difficult to segment the human subject from each depth image) which also lead to the lowest recognition accuracy.

\begin{figure}[tbp!]
    \centering
    \begin{subfigure}[t]{0.49\textwidth}
        \centering
        \includegraphics[width=\textwidth]{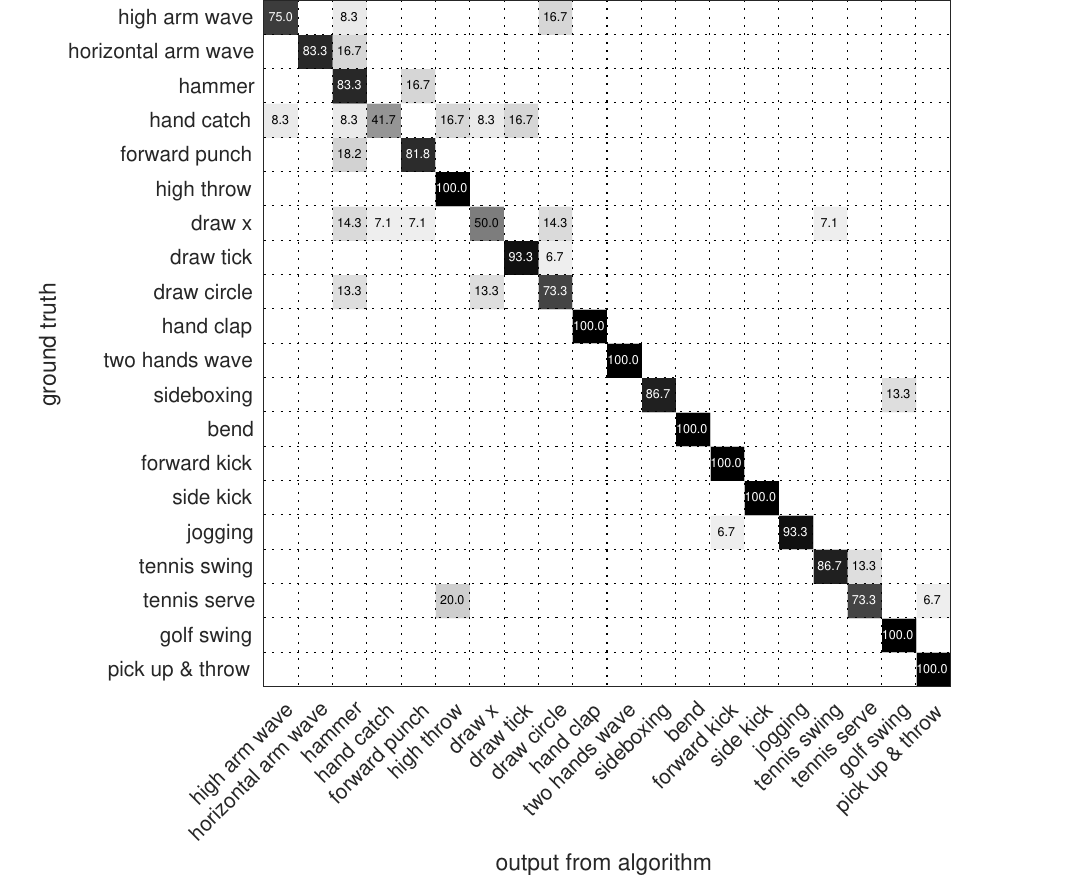}
        \caption{HDG-all features}
    \end{subfigure}
    \begin{subfigure}[t]{0.49\textwidth}
        \centering
        \includegraphics[width=\textwidth]{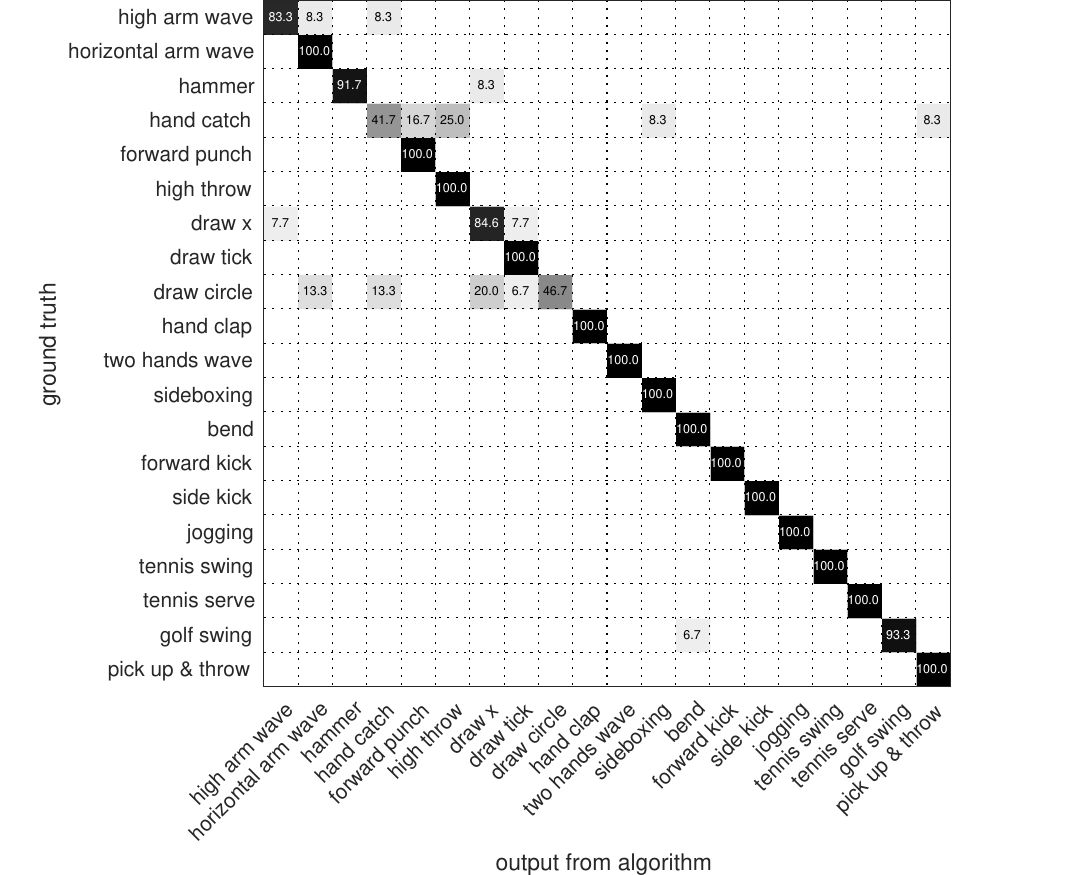}
        \caption{RBD-FTP representation}
    \end{subfigure}
    \caption{Confusion matrix for the MSRAction3D dataset when subjects \{2, 3, 5, 7, 9\} are used for training and the remaining subjects for testing.}
    \label{msrcomp}
\end{figure}

\subsection{MSRAction3D Dataset}

The confusion matrixes for the HDG algorithm with all feature vectors used and RBD-FTP representation~\cite{Vemulapalli2016} on the MSRAction3D dataset (when subjects \{2, 3, 5, 7, 9\} are used for training) are shown in Figure~\ref{msrcomp}. There are more confusion when using HDG than RBD-FTP representation~\cite{Vemulapalli2016}. We observe from the figure that the action {\it hand catch} and action {\it high throw} have higher confusion with each other in both methods as well as action {\it draw x} and action {\it draw circle} since the motion trajectories of these action pairs are similar. For HDG algorithm, the action {\it hammer} and action {\it forward punch} also have confusion because of similarities in motion. RBD-FTP representation~\cite{Vemulapalli2016} is a pure mathematic descriptor which encodes the skeleton rotation between pairs of body joints and it is robust to different human body size. Therefore, the recognition accuracy is higher than HDG-all features.

\subsection{3D Action Pairs Dataset}

\begin{figure}[tbp!]
    \centering
    \begin{subfigure}[t]{0.48\textwidth}
        \centering
        \includegraphics[width=\textwidth]{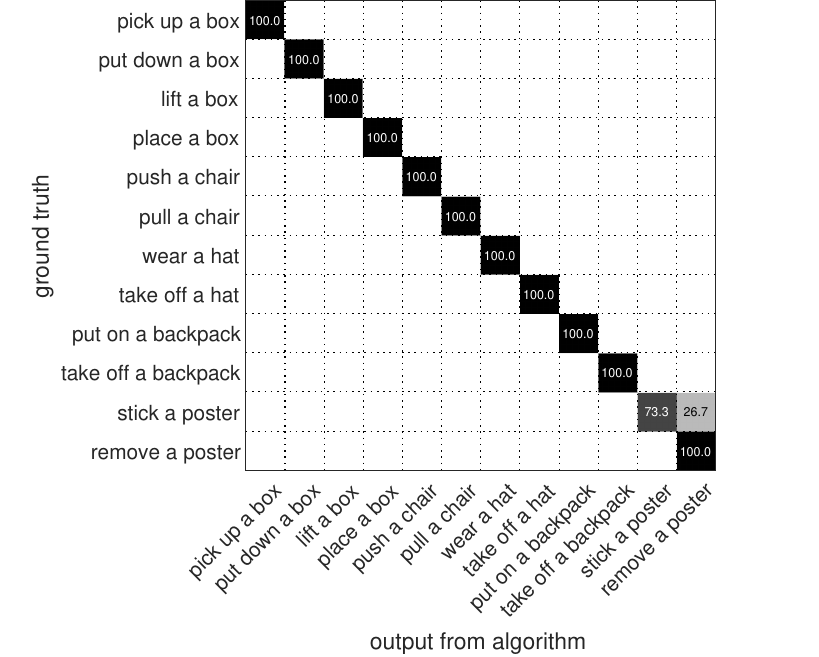}
        \caption{HDG-all features}
    \end{subfigure}
    \begin{subfigure}[t]{0.48\textwidth}
        \centering
        \includegraphics[width=\textwidth]{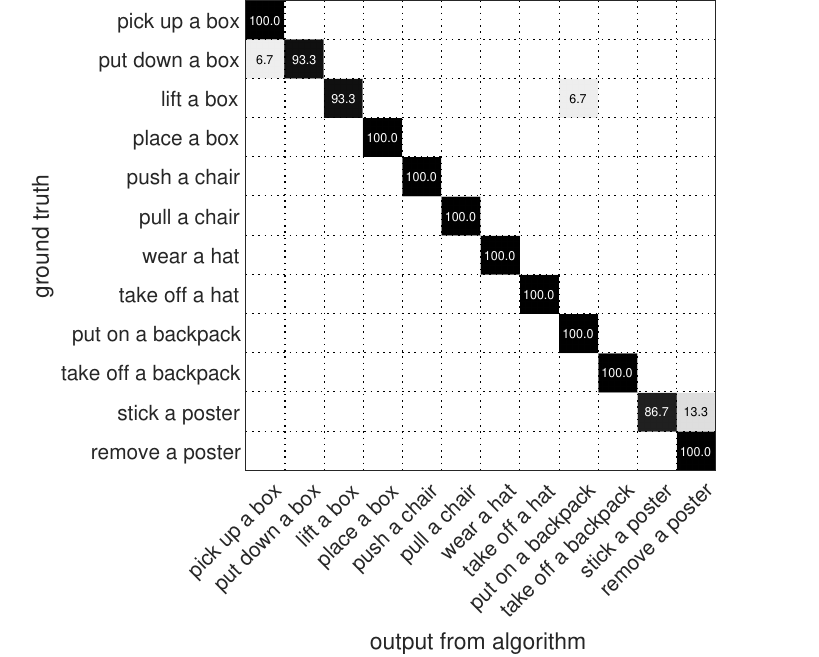}
        \caption{HON4D + Discriminative Density}
    \end{subfigure}
    \caption{Confusion matrix for 3D Action Pairs dataset when subjects \{1, 3, 5, 7, 9\} are used for training and the remaining subjects for testing.}
    \label{actionpairs}
\end{figure}

Figure~\ref{actionpairs} shows a comparison between HDG-all features and HON4D~\cite{Oreifej2013} on the 3D Action Pairs dataset. As shown in the figure, both algorithms perform equally well, but performing HDG-all features has less confusion than performing HON4D~\cite{Oreifej2013}. The main reason for this result is that HON4D~\cite{Oreifej2013} is extracted from depth sequence only, whereas HDG-all features is obtained from both depth sequence and skeleton data, and it contains more discriminative information than HON4D~\cite{Oreifej2013}. Both algorithms have confusion with action {\it stick a poster} and {\it remove a poster} due to the similar motions between two actions.

\begin{figure}[tbp!]
    \centering 
\begin{subfigure}{0.45\textwidth}
  \includegraphics[width=\linewidth]{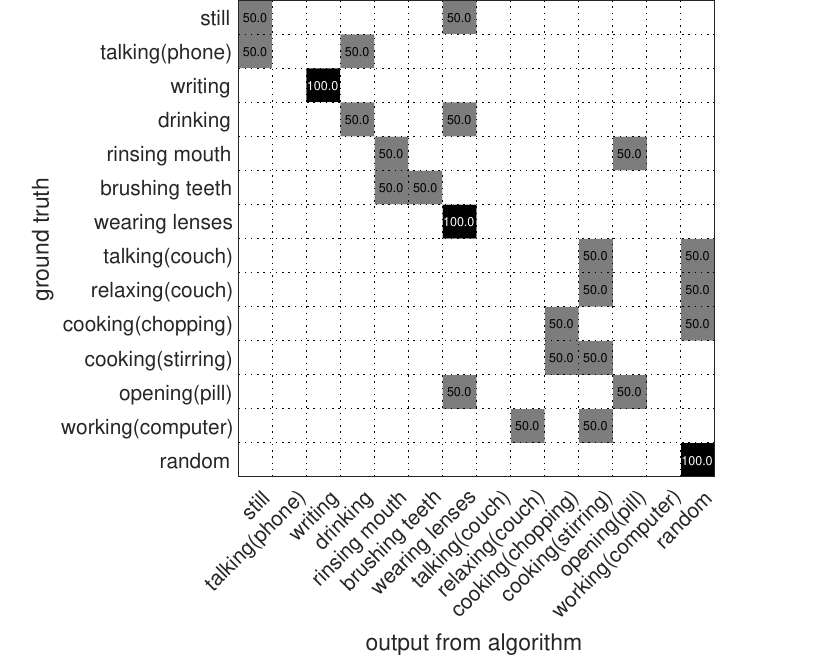}
  \caption{HDG (half training)}
  \label{hdg1}
\end{subfigure}\hfil 
\begin{subfigure}{0.45\textwidth}
  \includegraphics[width=\linewidth]{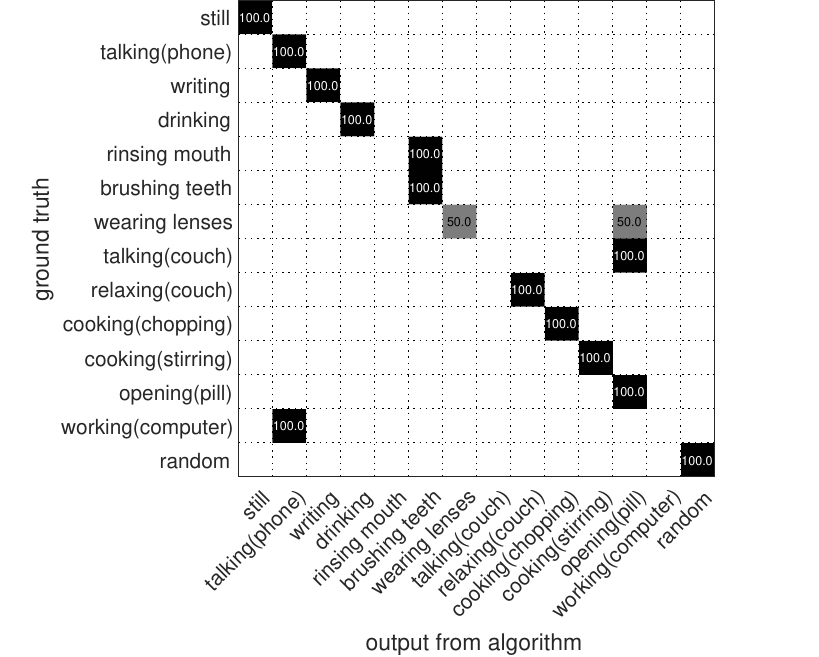}
  \caption{HDG ($3/4$ training)}
  \label{hdg2}
\end{subfigure}
\medskip
\begin{subfigure}{0.45\textwidth}
  \includegraphics[width=\linewidth]{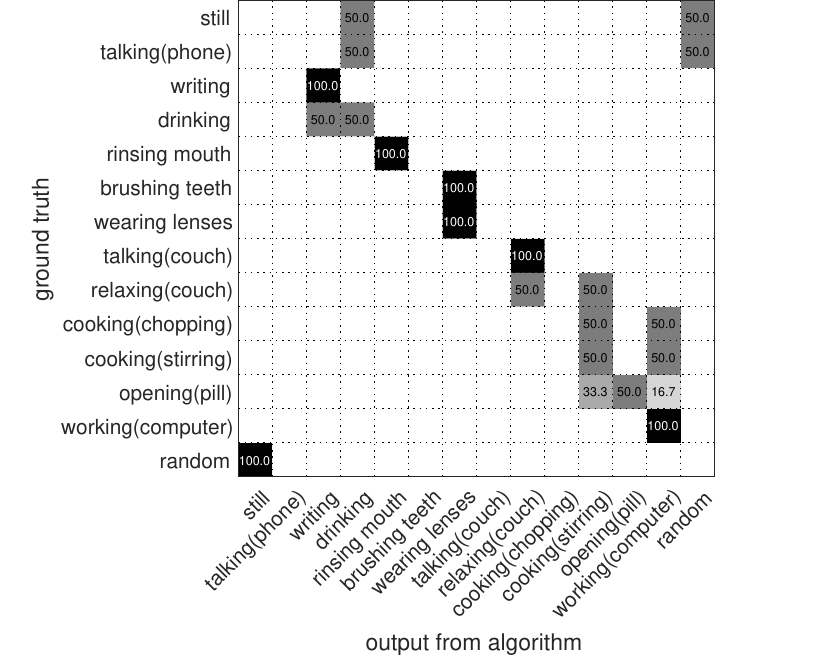}
  \caption{HOPC(half training)}
  \label{hopc1}
\end{subfigure}\hfil 
\begin{subfigure}{0.45\textwidth}
  \includegraphics[width=\linewidth]{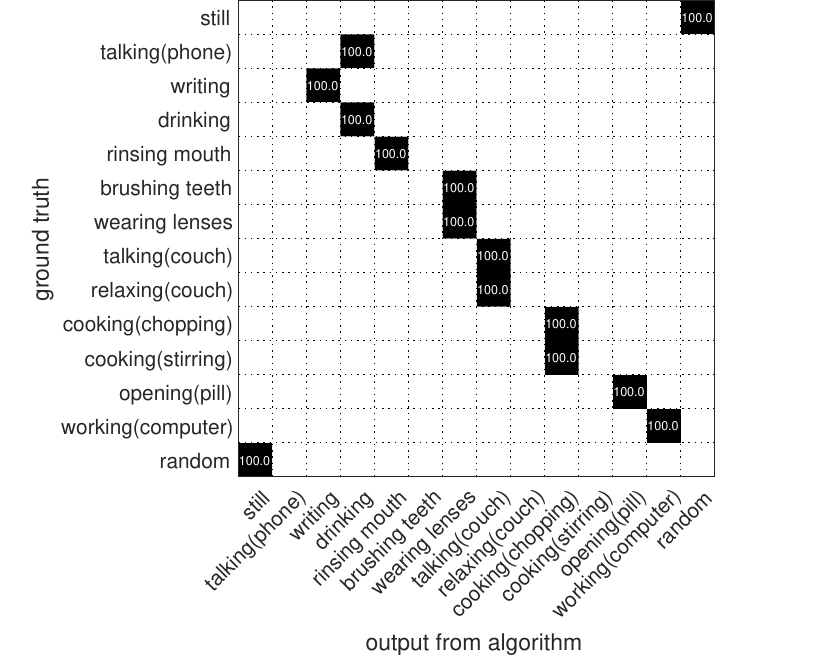}
  \caption{HOPC($3/4$ training)}
  \label{hopc2}
\end{subfigure}
\caption{Confusion matrixes for CAD60}
\label{cad60}
\end{figure}

\subsection{CAD-60 Dataset}

Figure~\ref{cad60} shows a comparison between the HDG and HOPC~\cite{Rahmani2016} algorithms. Since there are only 4 subjects in the CAD60 dataset and most actions are performed by one subject only once, the recognition accuracy is much lower when half of the subjects are used for training and the remaining half for testing. As can be seen in these 4 confusion matrixes, once more subjects ($3/4$ subjects for training) are used in training, the recognition accuracy increases a lot no matter which algorithm is used. However, when half of the subjects are used in training, the HOPC~\cite{Rahmani2016} performs worse than HDG in general. Some actions such as {\it brushing teeth} and {\it random} are totally confused when using HOPC~\cite{Rahmani2016}, even if more subjects are used in training. The reason for this is that HDG-all features contains features from both depth sequence and skeleton data, which is more reliable than HOPC~\cite{Rahmani2016}. However,  for HOPC~\cite{Rahmani2016}, the pointclouds are extracted from very limited number of depth sequences, even though it is a robust descriptor. If there are more samples (video sequences) collected in this dataset, HOPC~\cite{Rahmani2016} seems to perform better than HDG since this dataset contains a lot of background noise and HOPC~\cite{Rahmani2016} is more robust to noise.

\subsection{UWA3D Single View Dataset}

\begin{figure}[tbp!]
\begin{center}
\includegraphics[width=0.86\textwidth]{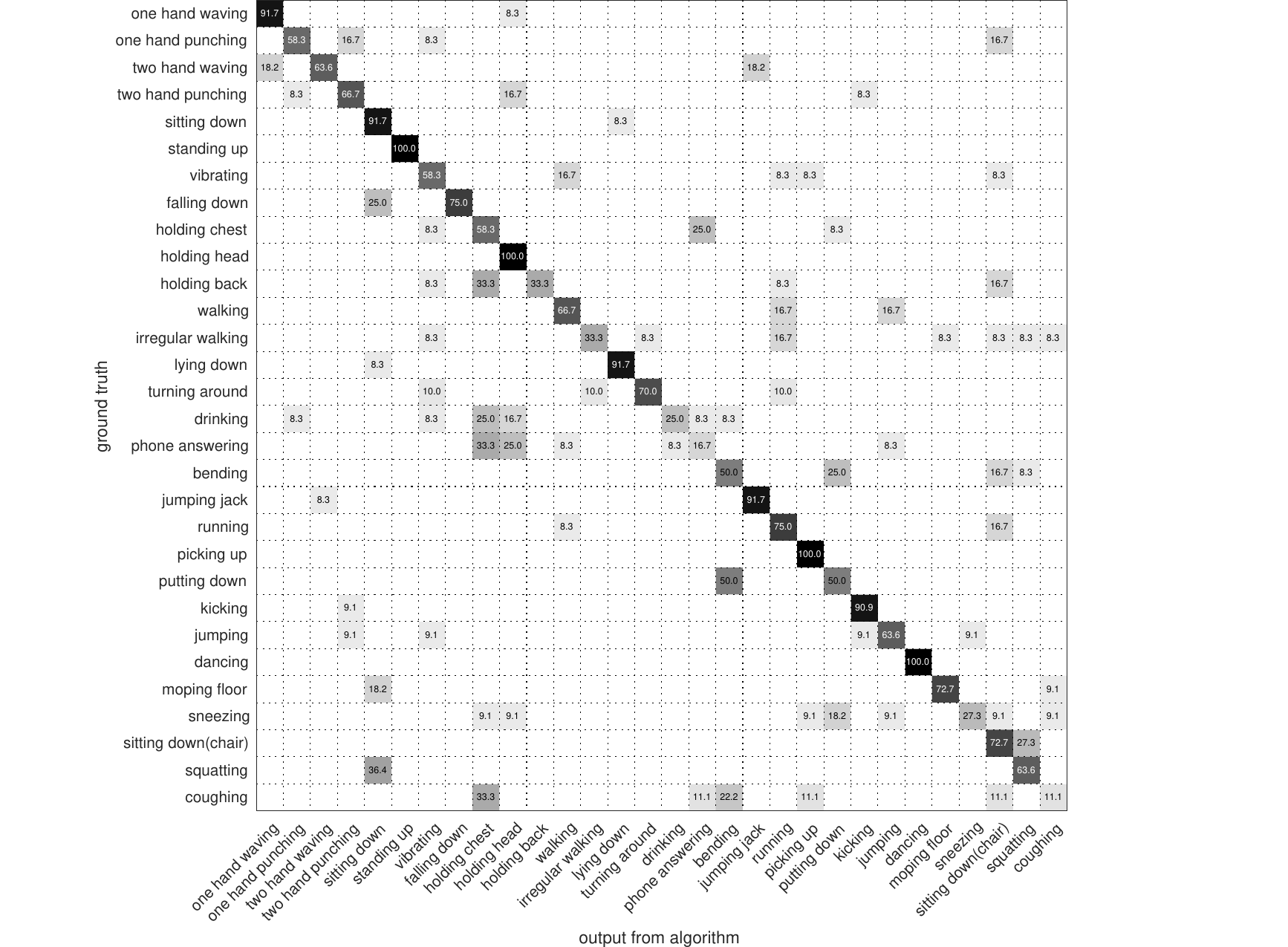}
\caption{Confusion matrix of the HDG algorithm with all features used on the UWA3D Single View dataset when subjects \{1, 3, 5, 7, 9\} are used for training and the remaining subjects for testing.}
\label{uwasingleview}
\end{center}
\end{figure}

The UWA3D Single View dataset is a little bit challenging since there are more actions in the dataset. This dataset contains 30 actions performed by 10 people in cluttered scenes with different human body size and different speed of performing actions. 

Figure~\ref{uwasingleview} shows a confusion matrix of HDG-all features on the UWA3D Single View dataset when subjects \{1, 3, 5, 7, 9\} are used for training and the remaining subjects for testing. Some actions such as {\it holding back}, {\it irregular walking}, {\it drinking}, {\it phone answering}, {\it sneezing} and {\it coughing} have the most confusion. This method cannot distinguish among actions such as {\it drinking}, {\it phone answering}, {\it holding chest} and {\it holding head} since the appearance and motion of these actions are very similar, which causes a confusion cluster shown in the confusion matrix.

\section{Results and Discussions for the UWA3D Multiview Activity Dataset}

The UWA3D Multiview Activity dataset is more challenging than other aforementioned four datasets since this dataset contains four different views (front, left, right and top view). To use this dataset for cross-view action recognition, two views of the samples are used for training and the remaining views for testing. For HOPC~\cite{Rahmani2016}, the constant temporal scale and spatial scale values given in \cite{RahmaniHOPC2014} are used in the experiments. Table~\ref{uwa3dmresults}~summarizes the results for cross-view action recognition, and the highest two recognition accuracy in each column have been highlighted.

\begin{table}[bp!]
\begin{center}
\caption{Comparison of action recognition accuracy (percentage) on the UWA3D Multiview dataset.}
\resizebox{\textwidth}{!}{\begin{tabular}{ l  c  c  c  c  c c  c  c  c  c c  c  c  }
\toprule
Training view & \multicolumn{2}{c}{$V_1$ \& $V_2$} & \multicolumn{2}{c}{$V_1$ \& $V_3$} & \multicolumn{2}{c}{$V_1$ \& $V_4$} & \multicolumn{2}{c}{$V_2$ \& $V_3$} & \multicolumn{2}{c}{$V_2$ \& $V_4$} & \multicolumn{2}{c}{$V_3$ \& $V_4$} & Mean\\
\hline
Testing view & $V_3$ & $V_4$ & $V_2$ & $V_4$ & $V_2$ & $V_3$ & $V_1$ & $V_4$ & $V_1$ & $V_3$ & $V_1$ & $V_2$ & {}\\
\hline
\hline
HON4D~\cite{Oreifej2013} & 31.1 & 23.0 & 21.9 & 10.0 & 36.6 & 32.6 & 47.0 & 22.7 & 36.6 & 16.5 & 41.4 & 26.8 & 28.9\\
\hline
HOPC~\cite{Rahmani2016} & 25.7 & 20.6 & 16.2 & 12.0 & 21.1 & 29.5 & 38.3 & 13.9 & 29.7 & 7.8 & 41.3 & 18.4 & 22.9\\
${\text{Holistic HOPC}^*}$ ~\cite{Rahmani2016} & 32.3 & 25.2 & 27.4 & 17.0 & 38.6 & 38.8 & 42.9 & 25.9 & 36.1 & 27.0 & 42.2 & 28.5 & 31.8\\
${\text{Local HOPC+STK-D}^*}$ ~\cite{Rahmani2016} & 52.7 & 51.8 & {\bf 59.0} & {\bf 57.5} & 42.8 & 44.2 & 58.1 & 38.4 & 63.2 & 43.8 & 66.3 & 48.0 & 52.2\\
\hline
${\text{RBD-logarithm map}}$ \cite{Vemulapalli2016} & 48.2 & 47.4 & 45.5 & 44.9 & 46.3 & 52.7 & 62.2 & 46.3 & 57.7 & 45.8 & 61.3 & 40.3 & 49.9\\ 
${\text{RBD-unwrapping while rolling}}$ \cite{Vemulapalli2016}& 50.4 & 45.7 & 44.0 & 44.5 & 40.8 & 49.6 & 57.4 & 44.4 & 57.6 & 47.4 & 59.2 & 40.8 & 48.5\\
${\text{RBD-FTP representation}}$ \cite{Vemulapalli2016}&54.9 & 55.9 &50.0 &54.9& 48.1 & 56.0 & 66.5 & {\bf 57.2} & 62.5 & 54.0 & 68.9 & 43.6 & 56.0\\ 
\hline
\hline
HDG-hod & 22.5 & 17.4 & 12.5 & 10.0 & 19.6 & 20.4 & 26.7 & 13.0 & 18.7 & 10.0 & 27.9 & 17.2 & 18.0\\
HDG-hodg & 26.9 & 34.2 & 20.3 & 18.6 & 34.7 & 26.7 & 41.0 & 29.2 & 29.4 & 11.8 & 40.7 & 28.8 & 28.5\\
HDG-jpd & 36.3 & 32.4 & 31.8 & 35.5 & 34.4 & 38.4 & 44.2 & 30.0 & 44.5 & 33.7 & 44.4 & 34.0 & 36.6\\
HDG-jmv & 57.2 & 59.3 & {\bf 59.3} & 54.3 & 56.8 & 50.6 & 63.4 & 52.4 & 65.7 & 53.7 & 67.7& 56.9 & 58.1\\
{HDG-hod+hodg} & 26.6 & 33.6 & 17.9 & 19.3 & 34.4 & 26.2 & 40.5 & 27.6 & 28.6 & 11.6 & 38.4 & 29.0 & 27.8\\
{HDG-jpd+jmv} & {\bf 61.0} & 61.8 & {\bf 59.3} & {\bf 56.0} & 60.0 & 57.4 & 68.8 & 54.2 & {\bf 71.1} & {\bf 57.2} & 69.7 & 59.0 & {\bf 61.3}\\
{HDG-hod+hodg+jpd} & 31.0 & 43.5 & 25.7 & 21.4 & 45.9 & 31.1 & 53.2 & 35.7 & 38.0 & 11.6 & 49.7 & 38.3 & 35.4\\
{HDG-hod+hodg+jmv} & 59.0 & {\bf 62.2} & 58.1 & 52.0 & 62.5 & 57.1 & 66.0 & 54.2 & 67.7 & 52.7 & {\bf 70.3} & {\bf 61.1} & 60.2\\
{HDG-hodg+jpd+jmv} & 58.2 & 61.8 & 54.8 & 47.6 & {\bf 63.5} & {\bf 58.7} & {\bf 69.0} & 52.3 & 64.9 & 47.1 & 67.2 & 59.4 & 58.7\\
{HDG-all features} & {\bf 60.9} & {\bf 64.3} & 57.9 & 54.6 & {\bf 62.6} & {\bf 59.2} & {\bf 68.9} & {\bf 55.8} & {\bf 69.8} & {\bf 55.2} & {\bf 71.8} & {\bf 62.6} & {\bf 61.9}\\
\bottomrule
\end{tabular}}
\label{uwa3dmresults}
\end{center}
{\textsuperscript{*}\footnotesize{This result is obtained from \cite{Rahmani2016} for comparison.}}
\end{table}

It can be seen in Table~\ref{uwa3dmresults}, the HDG-all features performs the best for cross-view action recognition (except that when view 1 and view 3 are used for training, view 2 or view 4 is used for testing and when view 2 and view 3 are used for training and view 4 is used for testing) compared to other state-of-the-art approaches. Moreover, HDG-all features outperforms Local HOPC+STK-D~\cite{Rahmani2016} which is the best results at that time by 9.7\% on average. Skeleton features (jpd and jmv) in HDG seems more robust than depth features (hod and hodg) as presented in Table~\ref{uwa3dmresults}, and HDG-jpd+jmv outperforms HDG-hod+hodg by 33.5\% on average. In addition, HDG-jpd+jmv performs better than Local HOPC+STK-D~\cite{Rahmani2016} in all experiments (except that when view 1 and view 3 are used for training and view 4 is used for testing) by 9.1\% on average, and even when only individual feature vector jmv is used, there is still 5.9\% higher recognition accuracy than its nearest competitor Local HOPC+STK-D~\cite{Rahmani2016}.

There is no evidence to show that RBD-unwrapping while rolling~\cite{Vemulapalli2016} performs better than RBD-logarithm map~\cite{Vemulapalli2016} for cross-view action recognition as shown in the table, and RBD-logarithm map~\cite{Vemulapalli2016} achieves even higher recognition accuracy than RBD-unwrapping while rolling~\cite{Vemulapalli2016}. RBD-FTP representation~\cite{Vemulapalli2016} achieves the highest recognition accuracy comparing with RBD-logarithm map~\cite{Vemulapalli2016} and RBD-unwrapping while rolling~\cite{Vemulapalli2016}, because it uses FTP to handle temporal misalignment and noise issues. However, RBD~\cite{Vemulapalli2016} is a mathematical descriptor which is computationally expensive and time-consuming comparing with other state-of-the-art approaches.

\begin{figure}[tbp!]
    \centering
    \begin{subfigure}[t]{0.46\textwidth}
        \centering
        \includegraphics[width=\textwidth]{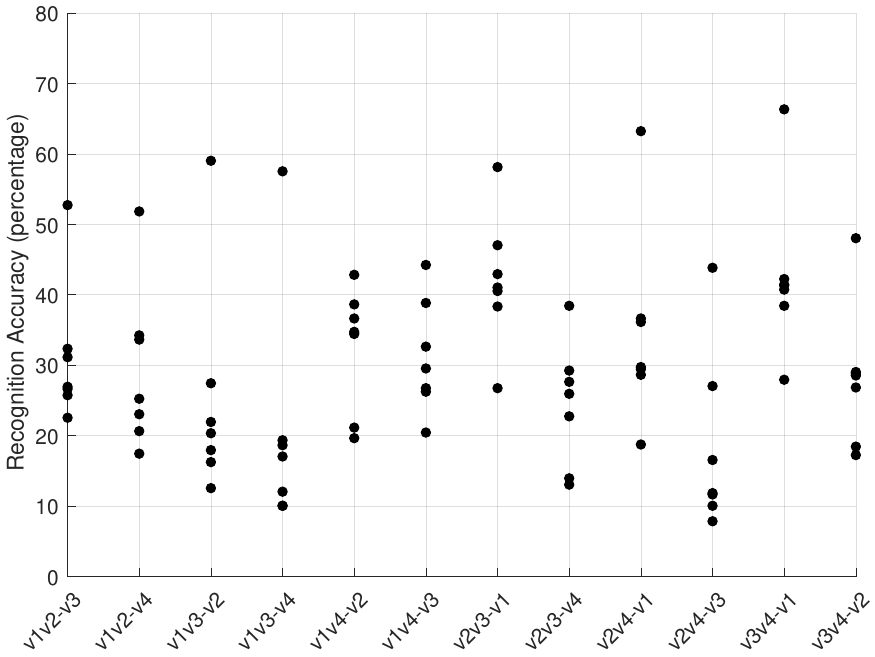}
        \caption{Depth-based algorithms.}
    \end{subfigure}
    \begin{subfigure}[t]{0.46\textwidth}
        \centering
        \includegraphics[width=\textwidth]{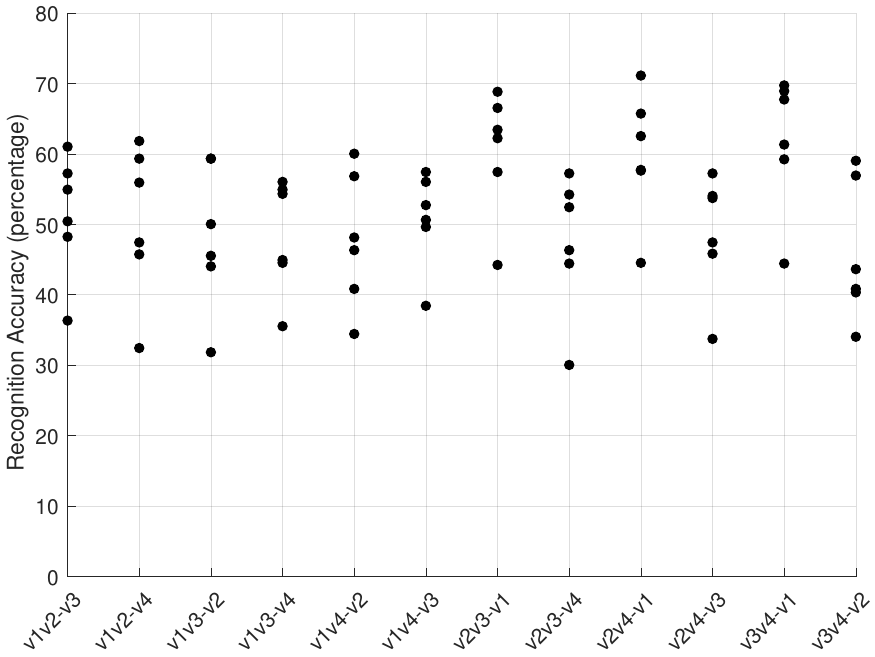}
        \caption{Skeleton-based algorithms.}
    \end{subfigure}
    \caption{Scatter plot of recognition accuracy for both depth-based and skeleton-based algorithms under different combinations of camera views in cross-view action recognition.}
    \label{scatterplot}
\end{figure}

Figure~\ref{scatterplot} is obtained based on the classification of algorithms given in table~\ref{uwa3dmresults}. We observe from this figure that skeleton-based algorithms are more robust than depth-based algorithms for cross-view action recognition. For example, some approaches which extract skeleton features from skeleton data such as RBD-FTP representation~\cite{Vemulapalli2016} and HDG-jpd+jmv perform better than the methods using depth features such as HON4D~\cite{Oreifej2013}, HDG-hod, HDG-hodg and HDG-hod+hodg. Depth feature such as Local HOPC+STK-D~\cite{Rahmani2016} can also increase the recognition accuracy since it encodes the discriminative information in an efficient way. However, skeleton features (jmv and jpd) can help to improve the recognition accuracy significantly for cross-view recognition, and they are much easier and faster to be computed than HOPC~\cite{Rahmani2016} and RBD~\cite{Vemulapalli2016}. Once the depth and skeleton features are combined together for cross-view action recognition, the descriptor would be more stable and robust, and the recognition accuracy should also increase (see Table~\ref{uwa3dmresults}). In addition, view 3 and view 4 (or view 2 and view 3 or view 2 and view 4) are used for training while view 1 is used for testing contributes to higher recognition accuracy. In contrast, when view 2 and view 4 are used for training and view 3 is used for testing (or view 1 and view 3 are used for training and view 4 is used for testing), the recognition accuracy is slightly lower. The reason for this result is that finding the correspondence between the learned classifier and testing views is difficult when cross-view action recognition strategy is applied on this dataset. 

\begin{figure}[tbp!]
\begin{center}
\includegraphics[width=0.76\textwidth]{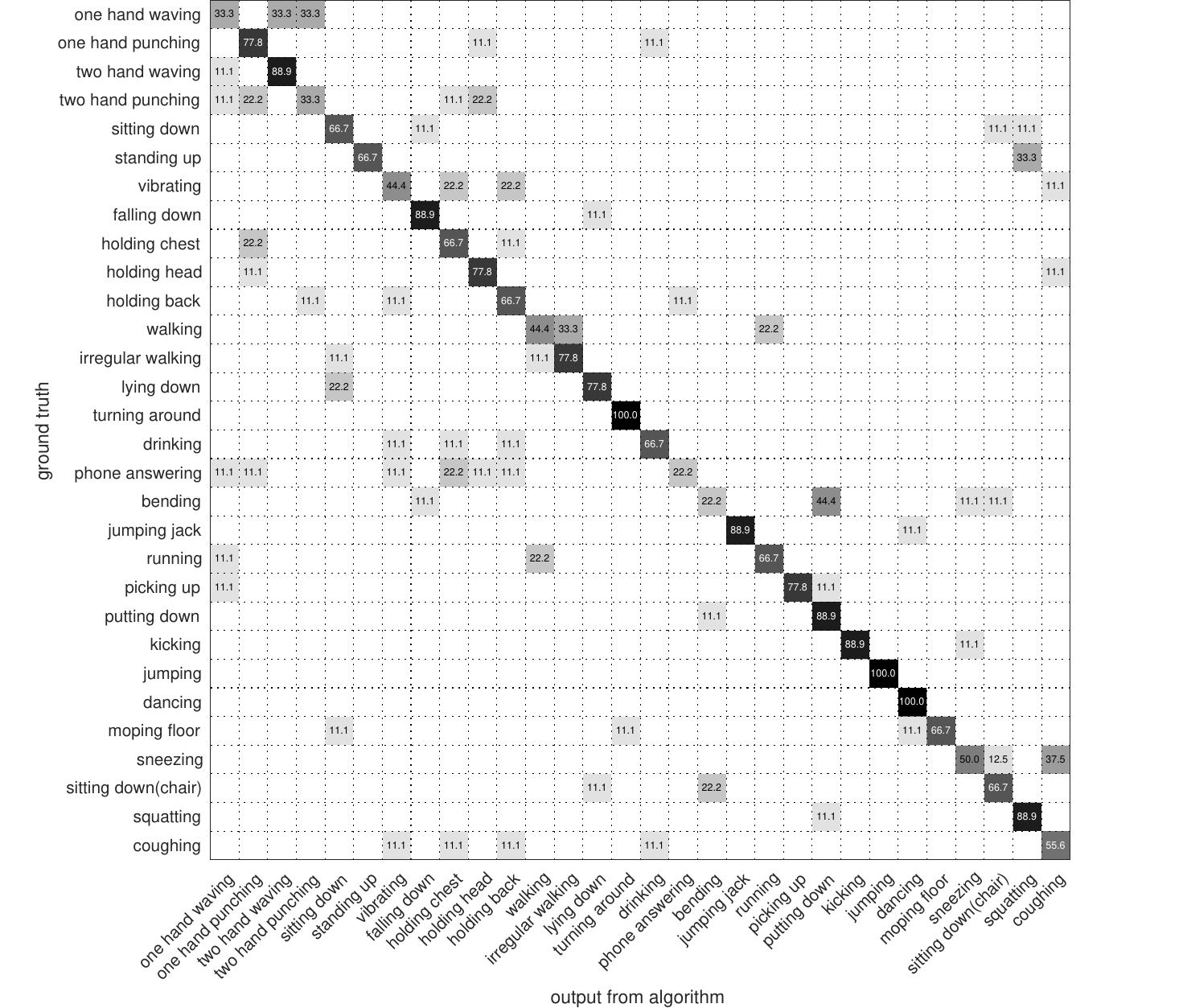}
\caption{Confusion matrix of HDG algorithm with all features used on UWA3D MultiView dataset when v3 and v4 are used for training and v1 is used for testing.}
\label{uwamultiviewv3v4v1}
\end{center}
\end{figure}

Figure~\ref{uwamultiviewv3v4v1} shows a confusion matrix of HDG-all features on the UWA3D MultiView dataset when view 3 and view 4 are used for training and view 1 is used for testing. As shown in this confusion matrix, the algorithm gets confused with some actions that have very similar motion trajectories and appearance. For example, {\it one hand waving} is similar to {\it two hand waving} and {\it two hand punching}; {\it walking} is very similar to {\it irregular walking}; {\it bending} and {\it putting down} have very similar appearance; {\it sneezing} and {\it coughing} are very similar actions.

\chapter{Conclusion and Future Work}

In this project we have analyzed and compared four state-of-the-art algorithms, namely HON4D~\cite{Oreifej2013}, HDG~\cite{Rahmani2014}, HOPC~\cite{Rahmani2016} and RBD~\cite{Vemulapalli2016}, on 5 benchmark datasets. We implemented and improved the HDG algorithm based on \cite{Rahmani2014}, and applied it to cross-view action recognition. Table~\ref{overallresults} summaries the performance of each algorithm for both single view and cross-view action recognition. As shown in the table, some algorithms perform better for single view action recognition but perform slightly worse in cross-view action recognition. The RBD-FTP representation~\cite{Vemulapalli2016}, on the other hand, performs better in both single view and cross-view action recognition, although it does not achieve the highest recognition accuracy for cross-view action recognition. 

\begin{table}[H]
\caption{Average recognition accuracy for both single view and multiview action recognition of all 4 algorithms.}
\begin{center}
\resizebox{0.8\textwidth}{!}{\begin{tabular}{| l | c | c |}
\hline
 Algorithms & ${\text {Single view action recognition}^{\dagger}}$ & ${\text {Cross-view action recognition}^{\ddagger}}$ \\ 
\hline
\hline
HON4D~\cite{Oreifej2013} & - & 28.9\\
\hline
HDG-all features & 70.0 & 61.9\\
\hline
HOPC~\cite{Rahmani2016} & 71.5 & 52.2\\
\hline
RBD-FTP~\cite{Vemulapalli2016} & 77.9 & 56.0\\
\hline
\end{tabular}}
\label{overallresults}
\end{center}
{\textsuperscript{$\dagger$}\footnotesize{For single view action recognition, the average recognition accuracy for each algorithm is averaged over the first 4 datasets.}}\\
{\textsuperscript{$\ddagger$}\footnotesize{For cross-view action recognition, the average recognition accuracy for each algorithm is averaged over 12 combinations of the training and testing views.}}
\end{table}

The experimental results for HDG show that skeleton features capture more discriminative information than depth features and they are more robust for cross-view action recognition. Our improved HDG-all features outperforms all other three methods for cross-view action recognition on the UWA3D Multiview Activity dataset. We also found that HOPC~\cite{Rahmani2016} is more robust to noise, human body size, and action speed variations although it is extracted from depth sequences only. Moreover, RBD~\cite{Vemulapalli2016} captures the relative 3D skeleton rotations between pairs of body joints which is robust to human body size; however, it is computationally expensive compared to other methods. 

The results suggest that it is important to capture enough discriminative information from human action videos efficiently, and the way we represent them is closely related to the performance of the algorithm. Our future work will focus on building a convolutional neural network (CNN) architecture to make it easier, faster and more robust than existing approaches in dealing with challenges such as noise, change of viewpoint and action speed, background clutter and occlusion. 

\pagebreak

\bibliographystyle{abbrvnat}

\bibliography{researchbib}

\end{document}